\documentclass[preprint,11pt]{elsarticle}

\usepackage{amssymb}
\usepackage{tabularx}
\usepackage[section]{placeins}
\usepackage{flafter}
\usepackage{amsmath}
\usepackage{booktabs}
\usepackage{multirow}
\usepackage{graphicx}
\usepackage{float}

\journal{IET Computer Vision}

\begin{document}

\begin{frontmatter}



\title{Sparse 3D Perception for Rose Harvesting Robots: A Two-Stage Approach Bridging Simulation and Real-World Applications}






\author[inst1]{Taha Samavati}
\author[inst1]{Mohsen Soryani\corref{correspondingauthor}} 
\ead{soryani@iust.ac.ir}
\cortext[correspondingauthor]{Corresponding author:}
\author[inst2]{Sina Mansouri}

\affiliation[inst1]{organization={School of Computer Engineering, Iran University of Science and Technology},
            addressline={Narmak}, 
            city={Tehran},
            state={Tehran},
            country={Iran}}

\affiliation[inst2]{organization={George Mason University Department of Computer Science},
            addressline={4400 University Drive}, 
            state={Virginia},
            country={United States}}

\begin{abstract}

The global demand for medicinal plants, such as Damask roses, has surged with population growth, yet labor-intensive harvesting remains a bottleneck for scalability. To address this, we propose a novel 3D perception pipeline tailored for flower-harvesting robots, focusing on sparse 3D localization of rose centers. Our two-stage algorithm first performs 2D point-based detection on stereo images, followed by depth estimation using a lightweight deep neural network. To overcome the challenge of scarce real-world labeled data, we introduce a photorealistic synthetic dataset generated via Blender, simulating a dynamic rose farm environment with precise 3D annotations. This approach minimizes manual labeling costs while enabling robust model training. We evaluate two depth estimation paradigms: a traditional triangulation-based method and our proposed deep learning framework. Results demonstrate the superiority of our method, achieving an F1 score of 95.6\% (synthetic) and 74.4\% (real) in 2D detection, with a depth estimation error of 3\% at a 2-meter range on synthetic data. The pipeline is optimized for computational efficiency, ensuring compatibility with resource-constrained robotic systems. By bridging the domain gap between synthetic and real-world data, this work advances agricultural automation for specialty crops, offering a scalable solution for precision harvesting.
\end{abstract}










\end{frontmatter}


\section{Introduction}
\begin{figure}[htpb]
    \centering
    \includegraphics{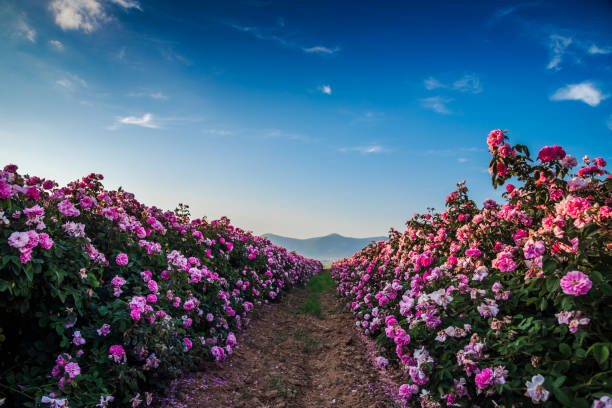}
    \caption{A sample damask rose farm \cite{ralij2022}.}
    \label{fig:real-farm-example}
\end{figure}

In recent decades, the increasing population and industrialization of human life have led to a significant rise in the demand for agricultural products. Consequently, stakeholders in the agricultural sector are striving to automate various processes of cultivation, maintenance, and harvesting to boost production capacity. Given that Damask rose holds a high cultivation area among medicinal plants in the Middle East, and its exports are projected to grow significantly in the coming years, developing a robot capable of detecting and estimating the depth of Damask roses to automate the harvesting process can enhance productivity and profitability. Figure \ref{fig:real-farm-example} shows a sample of Damask rose farms.

Recent advances in artificial intelligence and its widespread applications in computer vision have enabled depth estimation and 3D environmental understanding using two cameras or even a single camera. This can address the need for optimizing LiDAR and depth sensors, potentially reducing costs and ameliorating challenges such as predicting the depth of hidden parts of objects.

\begin{figure}[htpb]
    \centering
    \includegraphics[width=0.8\textwidth]{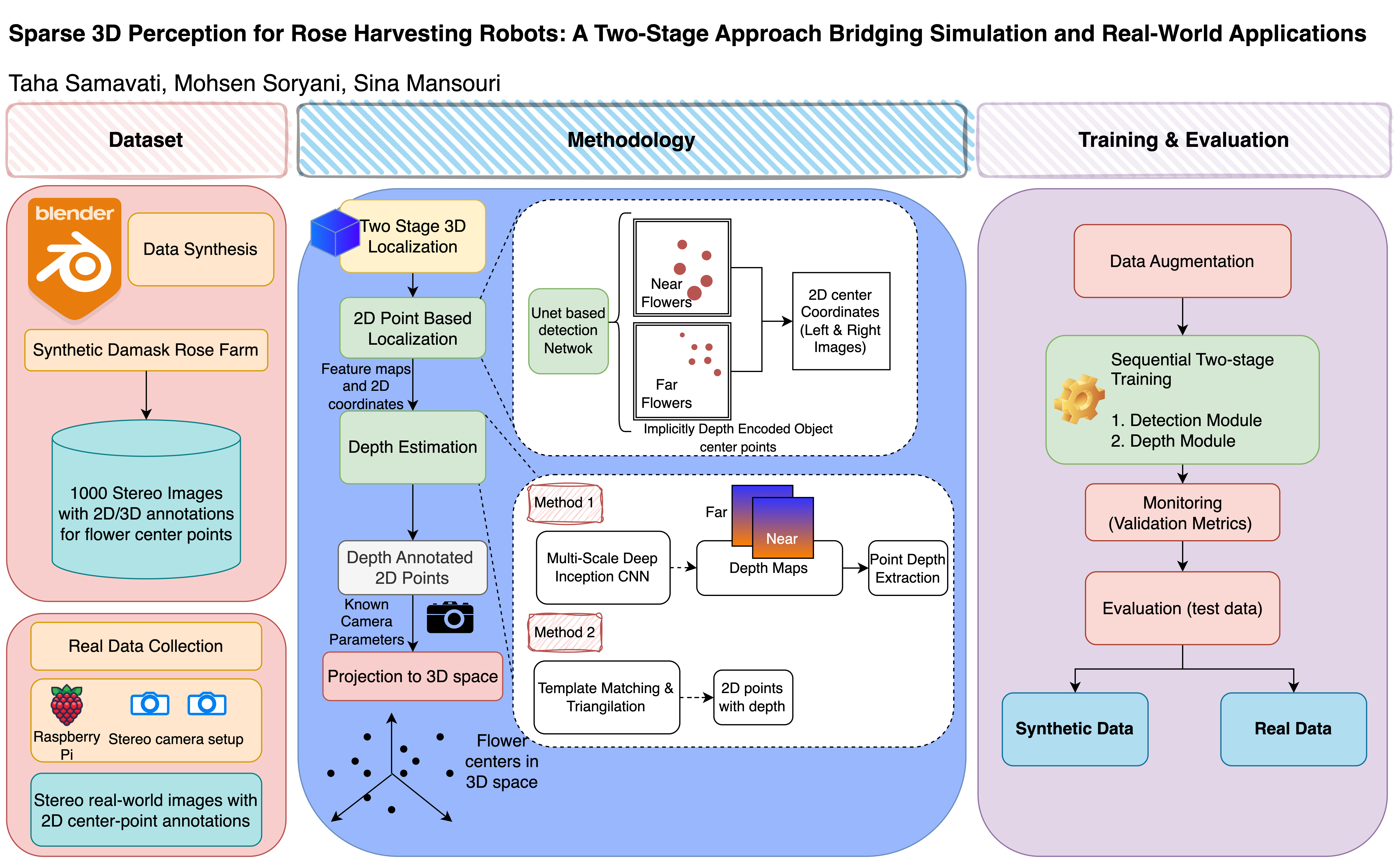}
    \caption{A visual summary of the study.}
    \label{fig:GA}
\end{figure}

This study explores 3D localization of Damask rose flower centers using stereo imaging and deep learning, proposing two approaches: an integrated method combining 2D detection with deep learning-based depth estimation, and a two-stage pipeline employing 2D object detection followed by template matching and geometric depth calculation. A monocular deep learning technique is also introduced to assess the viability of monocular depth estimation. To overcome the challenges of real-world 3D data acquisition, a synthetic dataset with 2D/3D annotations was generated, simulating various environmental conditions, while a limited real-world dataset with 2D labels was collected using a custom stereo setup. Experimental results demonstrate robust 2D localization accuracy and a mean depth estimation error of approximately 8 cm for stereo-based deep learning, achieving performance comparable to traditional template matching. Despite relying primarily on synthetic training data, the methods generalize to real-world scenarios, highlighting their practical potential for agricultural automation and precision horticulture applications. Figure \ref{fig:GA} presents a visual summary of this study.

\section{Related works}

\subsection{Object Detection}
Object detection is a critical area in computer vision and robotics, essential for robots to perceive and interact with their environment. Visual perception, particularly object detection, provides robots with the necessary information to understand and manipulate their surroundings. For example, in flower-picking tasks, 3D spatial information about flower locations is crucial for precise robotic arm movement. Recent years have seen significant research advancements in both 2D and 3D object detection, enhancing robots' ability to perform complex tasks in diverse environments. For example, YOLOv4 \cite{bochkovskiy2020yolov4} and VoxelNet \cite{zhou2018voxelnet} are capable of detecting and localizing objects present in a scene in 2D and 3D, respectively. 

Although many previous studies have utilized anchor-based and bounding box methods for object detection, recent investigations, such as those in \cite{duan2019centernet} and \cite{zhou2019objects}, have attracted significant research interest. In these studies, detection is conducted based on points instead of anchor-based object detection. This approach bears many similarities to segmentation algorithms. The advantages of such an approach compared to anchor-based methods include the elimination of the need to define multiple bounding boxes with variable aspect ratios, and the removal of the necessity to apply non-maximum suppression based on Intersection over Union (IoU). Additionally, there is no requirement for encoding and decoding labels during and after training, which simplifies the implementation significantly. This technique is also designed to more closely resemble human perceptual mechanisms, and it allows for dense prediction instead of sparse prediction.

\subsection{Depth Estimation}

Depth information is required to estimate the 3D coordinates of objects in a scene. Accurate depth estimation can be a challenging task. There are two primary approaches to obtaining depth: 1. direct measurement and 2. indirect measurement, which are explained below.

\subsubsection{Direct Measurement}
In this approach, 3D scene data is obtained using hardware. The hardware can be a depth sensor or a LiDAR. For example, research \cite{qi2018frustum} utilizes RGB-D camera data as input. Given an RGB image, this method first detects and localizes the object of interest in 2D using a convolutional neural network. At the same time, the depth map of the scene is converted into point clouds. Based on the camera frustum corresponding to the detected 2D bounding box, the 3D bounding box of the object is obtained. Research \cite{wang2015voting} utilizes LiDAR data and divides the 3D space of point cloud data into discrete cubic units. Each cell contains several points described by a fixed-dimensional vector. Subsequently, a 3D bounding box moves within this space, and a classifier determines the presence of an object in that region.

It's worth mentioning that the use of measurement devices also has its limitations. For example, LiDAR has a relatively high cost and is prone to errors when light is reflected from object surfaces. Additionally, since the power of the laser signal generator must be limited for safety reasons, it typically offers good accuracy only for short distances. Depth-sensing cameras, although they can have high accuracy up to certain distances, have errors in predicting the depth of edge points and are also very sensitive to motion blur.

\subsubsection{Indirect Measurement}

\textbf{Stereo Vision (Perspective View):} Given stereo images as input, research \cite{chen20173d} estimates a scene's depth map by using the algorithm presented in \cite{yamaguchi2014efficient}. Then, using the depth map and an energy function, it predicts the probable locations of objects after applying non-maximum suppression. Research \cite{li2019stereo} extends the Faster-RCNN algorithm \cite{ren2015faster} to stereo, detecting and locating cars in the scene in three dimensions.\\

\textbf{Monocular vision:} This approach has recently gained much attention from researchers. However, its performance is still inferior to other methods. This is because estimating depth from a single image is an ill-posed problem. Research studies \cite{chen2016monocular} and \cite{qin2019monogrnet} have utilized such an approach to estimate the depth of objects. For instance, research \cite{qin2019monogrnet} initially detects objects in the image in two dimensions using a convolutional neural network. Instead of predicting the entire scene depth map, a neural network is used to calculate only the depth of the detected objects' centers. It then estimates the coordinates of the bounding box corners in 3D space based on this information.\\

\textbf{Self-Supervised Methods:} Another recent research focus is the application of self-supervision in training depth prediction methods. In a stereo setup, an image from either the left or the right camera is fed to the network. The network then learns to reconstruct the opposite camera image by estimating a depth map for the input image. Then, using this prediction and knowing the camera parameters, the points from the first image are re-projected onto the second camera's image to estimate a 2D image of the second camera's view. The better the reconstruction quality of the second camera's image, the more accurate the predicted depth map. After predicting the depth map, 3D object coordinates can be extracted in a separate stage. Research \cite{garg2016unsupervised} utilizes this idea to estimate depth from a single image. During the training phase, depth labels are not necessary, and only stereo images are required. Research \cite{godard2017unsupervised} has enhanced the algorithm's performance by incorporating bidirectional (left-right) depth consistency criteria.

\subsection{Detection and Localization of Agricultural Products}
\textbf{2D localization}: Several studies have been conducted on 2D detection and localization of fruits and flowers, among which notable contributions include the research by \cite{wu2020using}, \cite{mazzia2020real}, and \cite{chu2021deep}. For instance, in the study by \cite{chu2021deep}, the 2D detection of apples in orchards was enhanced by adding a pruning branch to the Mask-RCNN model \cite{he2017mask}. This branch filters out non-apple pixels, leading to over 90\% detection accuracy. The model is trained using classification and IoU loss functions, with a binary cross-entropy loss for the pruning branch, enabling the network to effectively detect apples' 2D locations in images. 
Research \cite{hiary2018flower} explores 2D detection and classification of flowers using a convolutional autoencoder approach. The method employs VGG-16 \cite{simonyan2014very}, pretrained on ImageNet \cite{deng2009imagenet}, as the feature extractor. The proposed has two branches: one for pixel-wise segmentation of flowers, and another for flower type classification. After segmentation, the largest connected components are extracted, and their centroids are computed for 2D localization. It achieves average detection accuracies of 85.9\% on the Oxford102 \cite{nilsback2006visual} dataset, 82.2\% on the Oxford17 \cite{nilsback2006visual} dataset, and 81\% on the Zou-Nagy \cite{zou2004evaluation} dataset. Flower classification accuracy consistently exceeds 97\% on average. The Oxford17 dataset includes images and segmentation labels for 17 flower types, with 80 images per type under various angles and lighting. A larger dataset, Oxford102, features 102 flower types with 40 to 258 images each, but it does not include roses. Dataset \cite{zou2004evaluation} includes 612 images of 102 different types of flowers, with each category containing 6 images with a resolution of $240 \times 300$. 
Research \cite{kang2020fruit} performs 2D localization and semantic segmentation of fruits at multiple scales, which has been shown to improve results in various studies. The presented network has a branch for fruit localization and segmentation and another for tree branch segmentation. This aids in better understanding tree and fruit structures. The proposed model achieves a localization accuracy of 88.1\%. The accuracy of fruit and branch segmentation is reported as 85\% and 79\%, respectively. 
Researches \cite{wu2020using} and \cite{prakash2022flower} use Yolov4 to localizae flowers in 2D. For instance, research \cite{wu2020using} estimates 2D locations of apple blossom centers in images. The proposed algorithm finetunes a pruned Yolov4 model on collected real-world data. To achieve real-time execution on portable devices, the structure of the main model is pruned by removing channels of convolutional layers with minimal contribution to the output of each layer, reducing the number of parameters.
Research \cite{sun2021apple} explores 2D fruit blossom detection using a modified DeepLab-ResNet model \cite{chen2017deeplab}, pretrained on COCO-Stuff \cite{caesar2018coco}. The network is pruned to remove unrelated layers and fine-tuned. The method processes $321\times321$ image patches, applies segmentation refinement, and achieves F1-scores of 85\% for apple and pear, and 77\% for peach blossoms.
Research \cite{li2022real} presents a kiwi fruit localization algorithm using fine-tuned YoloV3 and YoloV4 models. The study used a dataset of 1451 labeled 2D points with depth information from a stereo camera system. YoloV4 outperformed YoloV3, achieving 2.4\% higher mAP on the test dataset and demonstrating 10\% greater generalization ability.

\textbf{3D localization}: Research \cite{stein2016image} focuses on 3D detection and localization of mango fruits using a robot equipped with a navigation system, camera, and LiDAR. The robot captures images along with displacement and time data as it moves, using the Faster-RCNN model for 2D mango detection. by appying triangulation, based on robot displacement and camera parameters, the fruits' 3D locations are determined. LiDAR data registers tree structures, allowing localized fruits to be accurately assigned to trees in the orchard. A dataset of 1500 mango tree images was used to train the model, achieving 88\% accuracy. Another approach involves using a five-channel input, including RGB, IR, and depth images from a Kinect v2 camera, which enhances localization accuracy to 94\%. Depth alignment also provides accessible depth information for detected objects. Given stereo images as input, research \cite{onishi2019automated} utilizes the Single Shot Multibox Detector (SSD) \cite{liu2016ssd} for 2D object detection in a single-stage, fully convolutional manner. Following the 2D detection phase, a point-cloud-based 3D reconstruction of the scene is obtained, allowing for the determination of the 3D coordinates of the pixels of interest.  Research \cite{kohan2011robotic} has employed a traditional image thresholding method for flower localization. It then estimates the depth of rose flowers. In this study, a robot with four degrees of freedom is responsible for capturing stereo imagery, computing the 3D coordinates of flower centers, and picking them. The main idea of the localization algorithm is to separate flower pixels from other image pixels based on color. After this stage, several morphological operations such as opening, erosion, and dilation are performed on the binary image obtained to identify all flowers, even overlapping ones. For depth estimation, considering that stereo images are rectified, searching is done on the scanline of the corresponding image. 
In research \cite{WANG2022106716}, the authors present the Apple 3D Network (A3N), a geometry-aware deep-learning model designed for robotic fruit (apple) harvesting. Building on existing research in object pose estimation using multi-source data, A3N combines RGB images and point clouds to enable end-to-end detection, instance segmentation, and grasping estimation of fruits using a RGB-D camera. The model utilizes a deep-learning detector for identifying Regions of Interest (ROIs) in RGB images, followed by bounding box regression on point clouds using PointNet \cite{qi2017pointnet} to predict optimal grasping angles for robotic arms. Furthermore, OctoMap \cite{hornung2013octomap} is incorporated to create an occupancy map, aiding the robot in accurately and safely navigating its environment. The A3N framework is evaluated with field data, showcasing its potential for enhancing robotic harvesting in orchards.
Research \cite{chen2024rapidstraw} tackles the challenge of identifying strawberries at various growth stages and determining optimal picking points for robotic harvesting. The authors propose an improved YOLO v8-Pose model \cite{Jocher_Ultralytics_YOLO_2023} combined with an RGB-D depth camera for fast and accurate detection. Enhancements include replacing Concat modules with BiFPN \cite{tan2020efficientdet} for better feature fusion, using the MobileViTv3 \cite{wadekar2022mobilevitv3} framework for improved contextual feature extraction, and switching from CIoU to SIoU loss function for faster convergence. These improvements led to a 97.85\% mAP-kp, a 5.49\% increase over the initial model. The method also accurately localizes picking points in 3D space, with a mean absolute error of 0.63 cm and a mean absolute percentage error of 1.16\%. 
Research \cite{JANG2024108961} tackles the challenges of accurately estimating tomato position and pose for efficient and safe robotic harvesting in smart farm environments. The authors propose a method that uses information from tomato sepals to estimate the fruit's orientation. A YOLOv8 model is trained to detect and segment both tomato bodies and sepals, utilizing data augmentation techniques for robust training. The method matches the segmented body and sepal using IoU scores and the Hungarian algorithm, then computes the pose by analyzing point clouds obtained from RGB-D data. Experimental results show high accuracy, with AP(50) scores of 94.7 for sepals and 96.3 for tomatoes, and a mean pose estimation error of 6.79 ± 3.18 degrees.

\section{Data Collection}
Deep neural networks require substantial amounts of data for effective performance. However, there is a lack of openly available datasets for locating flowers in 3D space. To address this, we created a synthetic dataset with precise 2D and 3D flower coordinates. Additionally, a complementary real-world dataset was curated to evaluate algorithm performance. The following sections detail the methods used to gather these datasets.

\subsection{Synthetic Data}
The cost-effectiveness of synthetic data generation has led many researchers and industry professionals to adopt this approach. It reduces costs, saves time, and allows for the creation of diverse, customizable environments, which are often difficult to replicate in reality. Deep learning models trained on synthetic data show competitive performance, which can be further enhanced by fine-tuning with a small amount of real data.

\begin{figure}[H]
    \centering
    \includegraphics[width=0.8\linewidth]{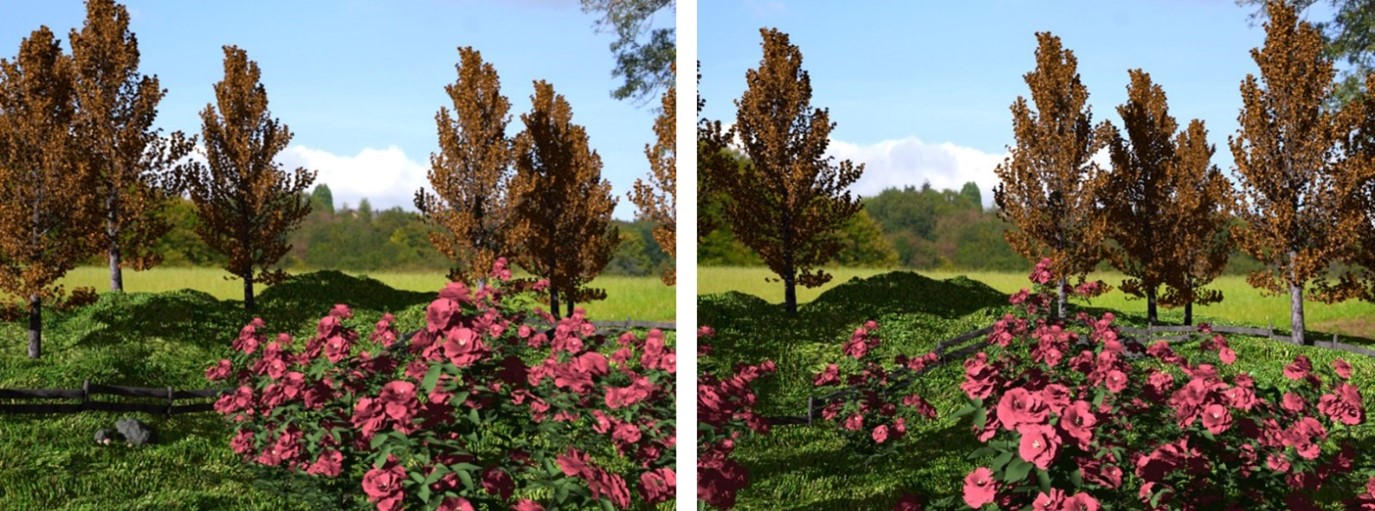}
    \caption{Two samples of synthetic images.}
    \label{fig:synthetic-farm-example}
\end{figure}

In this study, a realistic environment of a rose farm was created and simulated using Blender\cite{Blender}, a 3D modeling software. In this environment, a large number of rose bushes were placed in a farm setting. Figure \ref{fig:synthetic-farm-example} illustrates two samples of the synthetic images obtained from this simulation.

Each data sample includes stereo RGB images, the 2D location of the flower centers in the images, the 3D location of the flower centers, and the corresponding depth maps as labels. Ground truth heatmaps were generated by convolving 2D flower center coordinates with adaptive Gaussian kernels, where the spatial extent (sigma) of each kernel was inversely proportional to the flower’s depth. This encourages the model to learn sharper localization for nearer flowers.

The camera configuration and characteristics are provided in Table \ref{tab:stereo_parameters}.

\begin{table}[H]
\centering
\caption{The Parameters of the stereo camera system.}
\label{tab:stereo_parameters}
\begin{tabular}{|c|c|}
\hline
\textbf{Parameter} & \textbf{Value} \\ \hline
Focal length & 26 mm \\ \hline
Pixel Size & 0.325 mm \\ \hline
Stereo System Type & Parallel in the x-direction \\ \hline
Baseline & 65 mm \\ \hline
\end{tabular}
\end{table}

The dataset consists of 1000 samples. These data were generated by placing cameras at various locations. The cameras captured images of the flowers from various distances and angles, while simultaneously recording both 2D and 3D coordinates of the flowers' centers.

\subsection{Real Dataset}
\begin{figure}[H]
    \centering
    \includegraphics[width=0.8\linewidth]{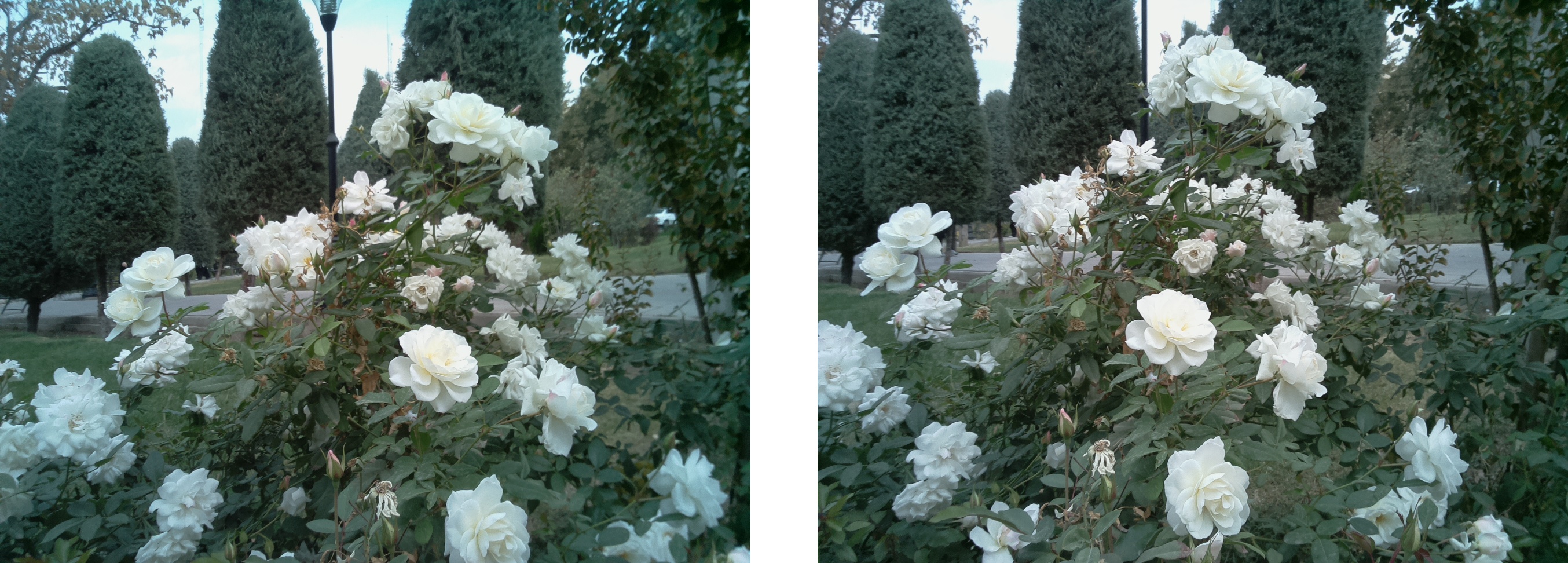}
    \caption{Two samples of rose images collected by a stereo camera. The left and right images correspond to the left and right cameras, respectively.}
    \label{fig:sample-real-data}
\end{figure}

Due to the necessity of evaluating the performance of the developed model with real data, a dataset of roses and similar flowers was collected. Due to constraints, the depth of flower centers was not recorded. To collect the data, two cameras with identical specifications were purchased. These two cameras were set up using two Raspberry Pi devices and placed at a specified distance from each other. By setting up software on a mobile phone, the capability to send commands to both devices was facilitated. After sending commands to the devices, they simultaneously capture images. In Figure \ref{fig:sample-real-data}, two different samples of data captured by the device are displayed. For each sample, stereo images were captured and 2D labeling was performed.


\begin{figure}[H]
    \centering
    \includegraphics[width=0.5\linewidth]{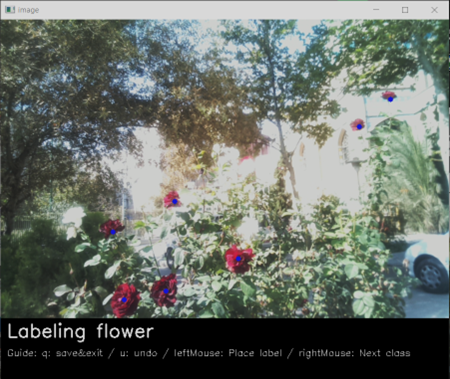}
    \caption{Demonstration of the labeling process of real data.}
    \label{fig:2D-labeling-software}
\end{figure}

A Python program was written for labeling the 2D coordinates of flower centers. Figure \ref{fig:2D-labeling-software} illustrates the labeling process in this software.

\subsection{Data Preparation}
Synthetic data serves as the training source for the proposed deep neural networks, as detailed in the subsequent section. In robotic flower grasping, precise localization of distant flower centers is non-essential, as the robotic arm prioritizes grasping nearby blooms. To reflect this operational constraint, flower centers are classified into near or distant categories using a distance threshold $\tau=2m$ from the camera. This binary classification is encoded within a three-channel heatmap framework: Channel 1 (near flowers), Channel 2 (distant flowers), and Channel 3 (background). Spatially adaptive Gaussian activations, modulated by depth information, ensure precise localization of near flowers while suppressing distant candidates. Stereo image pairs are processed to generate corresponding heatmap-depth tuples for both left and right camera perspectives, enabling robust correspondence estimation. To optimize training stability, heatmaps were normalized to the range [0,1], and depth values were scaled by a factor of $\alpha=\frac{1}{8}$, mitigating numerical instability during gradient-based optimization.

\subsubsection{Data Splitting}
\begin{figure}[H]
    \centering
    \includegraphics[width=0.9\textwidth]{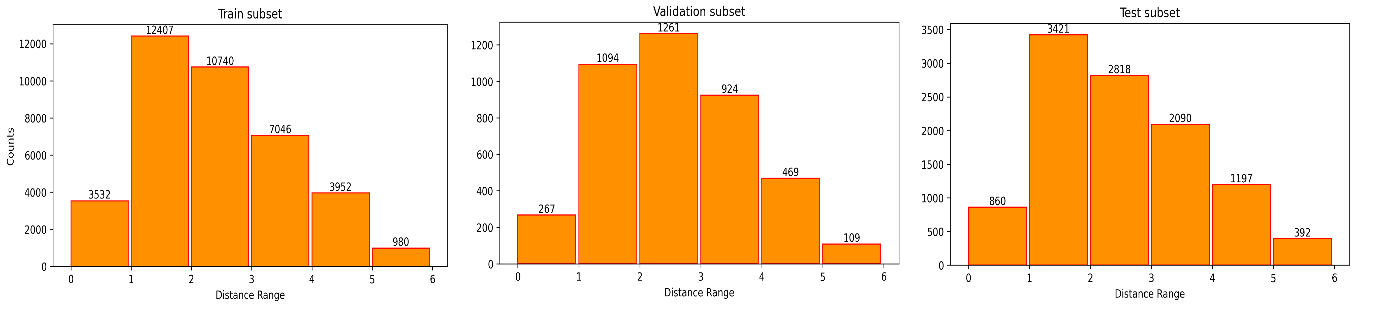}
    \caption{The distribution of distances between the left camera's center and flower centers (depth) across three dataset subsets. The horizontal axis represents distances in meters, while the vertical axis indicates the number of flowers within each distance range.}
    \label{fig:distance-distributions}
\end{figure}

The synthetic dataset was split into three parts: training, validation, and test sets, with respective ratios of 70\%, 15\%, and 15\%. The validation dataset is used to monitor the model's performance during training and prevent overfitting. The histograms in Figure \ref{fig:distance-distributions} illustrate the distribution of distances between flower centers and the camera in each of these subsets. As observed, the ratios are reasonably consistent across all subsets.

\subsubsection{Data Augmentation}
To mitigate overfitting and improve generalization, data augmentation artificially diversifies training datasets. This study employs standard image transformations—random brightness/contrast adjustments and compression artifacts—alongside environmental simulations (rain, sunlight effects) to address data scarcity and enhance real-world robustness. Augmentations were applied exclusively to training data, with parameters tuned to preserve stereo correspondence (e.g., rain/sunflare effects were applied identically to both left and right images).

\section{Proposed Method}
\subsection{Monocular 3D localization}
\begin{figure}[H]
\centering
\includegraphics[width=0.7\textwidth]{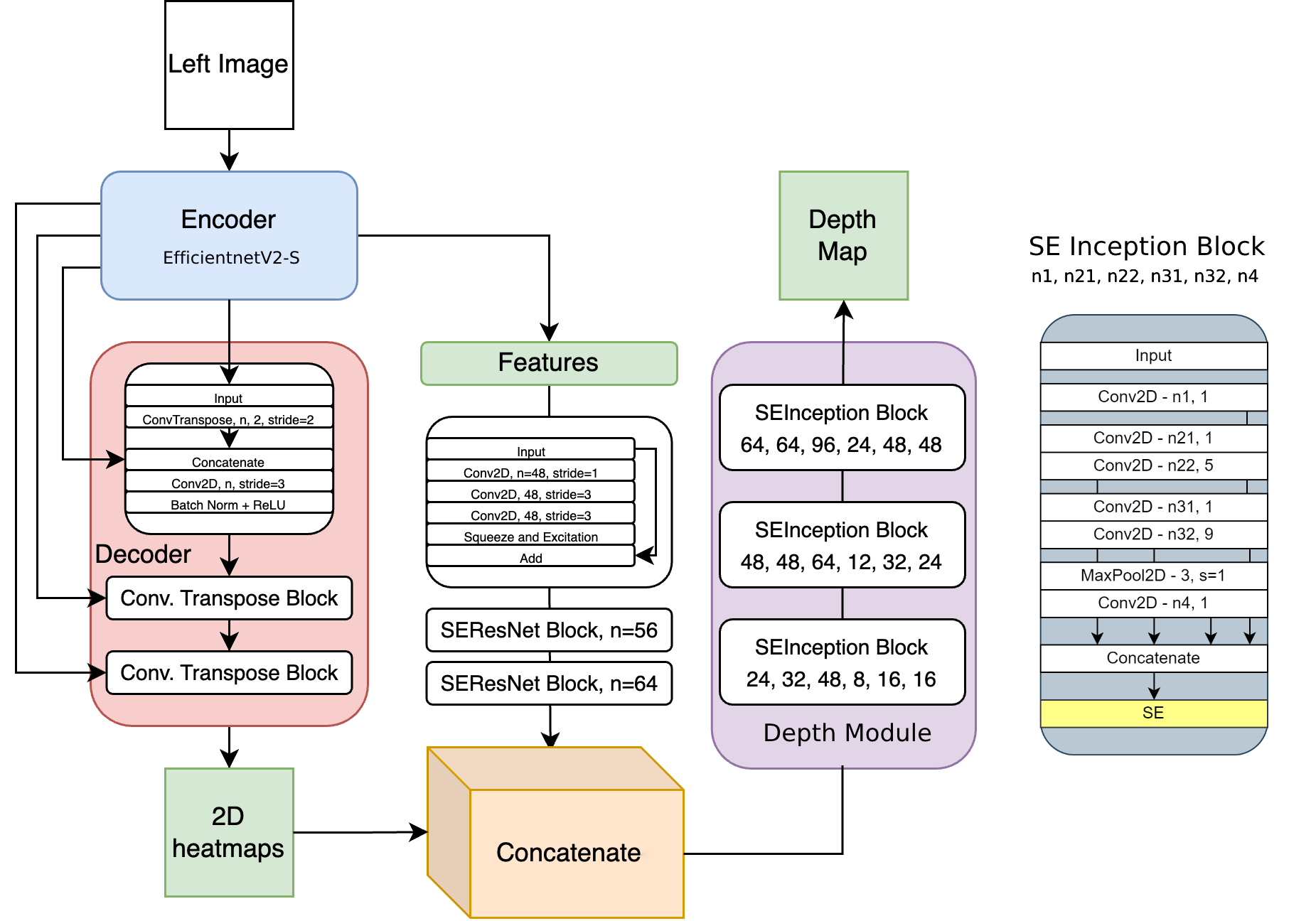}
\caption{An overview of the proposed monocular deep learning based model for 3D localization of flower centers.}
\label{fig:single-view-structure}
\end{figure}

Figure \ref{fig:single-view-structure} illustrates the architecture of the proposed neural network. This network consists of two main parts: 2D localization and depth estimation. After the flower centers were detected and localized in 2D by the first module, the results, along with the intermediate features of the images, are fed to the depth estimation module to estimate the depth of the flower centers.

\subsubsection{2D Flower Localization}\label{subsubsec:2dflowerloc}
In this research, a point-based detection approach is utilized for the task of 2D detection and localization instead of an anchor-based approach. This localization network has a U-Net-like architecture \cite{ronneberger2015u}. This network is widely used in many applications, especially in semantic segmentation. The encoder is an EfficientNetV2-S \cite{tan2021efficientnetv2} with pre-trained weights on the ImageNet dataset. The decoder transforms the feature vector of the encoder to a higher spatial dimension in three steps. In each step, a transposed convolution operation is performed on the previous stage until a heatmap with identical dimensions as the input image results. The skip-connections between the encoder and the decoder help preserve spatial information by concatenating corresponding feature maps from the encoder to the decoder, thereby enhancing the accuracy of the heatmap. This design allows the network to leverage both the high-level abstract features and the detailed spatial information, leading to more precise localization in the 2D detection task. The resulting heatmap has three channels, each of which corresponds to the probability map of near flower centers, distant flower centers, and the probability map of the background, respectively. 

\subsubsection{Monocular Depth estimation}
In the depth estimation phase, the input image features pass through three SEResNet blocks and are further concatenated with 2D heatmaps of the localization output. These results are then fed into a stack of three SE-enhanced Inception blocks \cite{szegedy2015going} and a dense depth estimation of the scene is obtained. The Inception block is used to apply multiple filters with various spatial dimensions simultaneously. This approach conceptually enables the network to have multiple fields of view, allowing it to better estimate the depth of both near and distant points by integrating this diverse spatial information.

To improve convergence during training, a custom sigmoid activation is applied to the depth estimation output, defined as:

\begin{equation}
    \text{custom\_sigmoid}(x) = \sigma\left(\frac{x - 5}{2}\right),
\end{equation} where \(\sigma\) is the standard sigmoid function. This transformation ensures that depth predictions are constrained to a reasonable range, facilitating stable gradient updates. During evaluation, the inverse transformation:  

\begin{equation}
\text{pred} = -2 \log\left(\frac{1}{\text{pred}} - 1\right) + 5
\end{equation}

is applied to recover the original depth values, ensuring consistency between training and inference.

\subsubsection{Loss Function}
There are two loss functions defined; one for localization and another for depth estimation. The 2D localization loss which is a weighted multi-class cross-entropy \cite{mao2023cross} and is defined as follows:

\begin{equation}
L_{CCE} = \sum_{i \in C} w_i y_i \log(p_i), \quad C \in \text{\{background, near flower, distant flower\}}
\end{equation}

Where $p_i$ is the probability of a pixel belonging to class $i$, $y_i$ is the binary value of belonging to class $i$, and $w_i$ is the weight assigned to each class. Here, three classes are considered for each pixel in the input image. These classes include background, near flower center, and distant flower center. The class weights are considered based on the importance of each class, with near flower having the highest weight and the background class having the lowest.

\begin{figure}[H]
\centering
\includegraphics[width=0.4\textwidth]{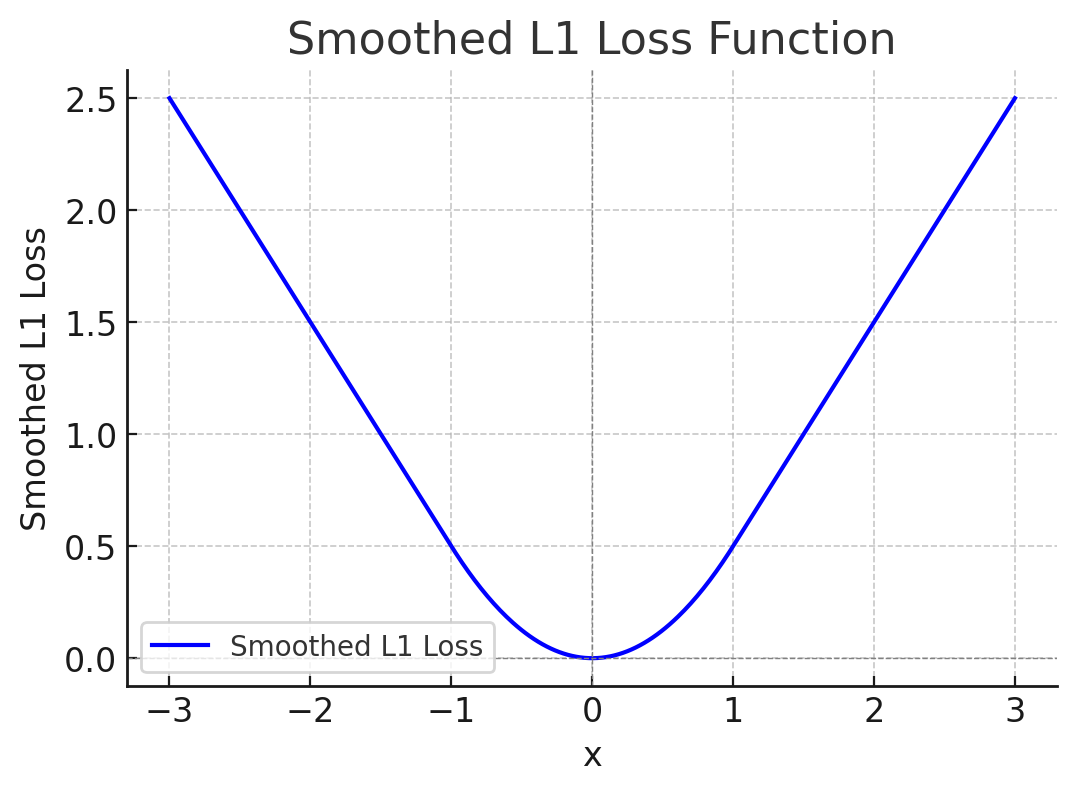}
\caption{Plot of the loss function for depth estimation.}
\label{fig:smoothed-l1}
\end{figure}

A smoothed L1 loss \cite{girshick2015fast} is used as an objective function to train the depth estimation module. It should be noted that supervision is only applied to the flower center pixels. This loss function is defined as follows:

\begin{equation}
L_{Depth} = \frac{1}{N} \sum_{i=1}^N L_{Smoothed L1}
\end{equation}
\begin{equation}
L_{Smoothed L1} =
\begin{cases}
\frac{1}{2} x^2 & \text{if  $\left|x\right|<1$}\\
\left|x\right| - 0.5 & \text{otherwise},
\end{cases}
\end{equation}

where $N$ is the number of samples in a batch. A key advantage of this loss function is its differentiability at the origin, unlike the common L1 loss function. It combines the benefits of L1 loss (constant gradient for large errors) with those of L2 loss (reduced oscillation and easier convergence for small errors). This loss function is plotted in Figure \ref{fig:smoothed-l1}.

The final loss function is obtained by summing the two mentioned parts and is as follows:

\begin{equation}
L_{total} = L_{CCE} + L_{Depth}
\end{equation}

\subsection{Stereo 3D localization}
This section outlines the algorithm for 3D damask rose localization using stereo images. Two approaches were developed and compared: (1) 2D flower center detection in the left image followed by depth estimation via triangulation with known camera parameters, and (2) a dual deep learning framework combining 2D localization (identical to the monocular method) with a dedicated depth estimation module. The 2D localization structure mirrors that of the monocular method (see Section \ref{subsubsec:2dflowerloc} for details), while subsequent sections elaborate on each 3D approach.

\subsubsection{Depth Estimation using template matching and epipolar geometry}
After localizing flower centers in the left image using the 2D detection module, depth estimation proceeds by extracting these centers from the probability maps via connected components analysis. Correspondence between stereo images is constrained to horizontal scanlines due to the rectified camera geometry, enabling efficient search for matching points in the right image. A $32\times32$ patch centered on each left-image point is compared against a horizontally bounded strip in the right image, where the strip’s width is limited by the dataset’s maximum depth to optimize computational efficiency.

The matching process employs normalized cross-correlation (NCCorr) \cite{opencvtemplatematching} to identify correspondence. The peak correlation response within the search strip determines the right-image flower center. The formula is defined as in Eq. \ref{eq:NCCorr}:

\begin{equation}
R(x,y) = \frac{\sum_{x',y'} (T(x',y') \cdot I(x+x',y+y'))}{\sqrt{\sum_{x',y'}T(x',y')^2 \cdot \sum_{x',y'} I(x+x',y+y')^2}}
\label{eq:NCCorr}
\end{equation}

where, $T$ represents the template and $I$ represents the image. The formula for the normalized correlation coefficient method is mostly the same \cite{opencvtemplatematching}. The only difference is that both the template and image pixel values are normalized before the response is calculated. Its abbreviation in this research is NCCoef. If we define the new template and image values as $T'$ and $I'$, they are defined as:

\begin{align}
T'(x',y') &= T(x',y') - \frac{1}{w \cdot h} \cdot \sum_{x'',y''} T(x'',y'')\\
I'(x,y) &= I(x,y) - \frac{1}{w \cdot h} \cdot \sum_{x',y'} I(x+x',y+y')
\end{align}

\begin{figure}[H]
\centering
\includegraphics[width=0.8\textwidth]{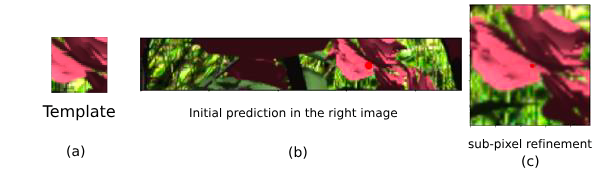}
\caption{An overview of the template matching process from the left camera image to the right camera image. (a) shows the template. (b) The search area in the right camera image. (c) Illustrates the increased dimensions around the initial estimate to provide a sub-pixel estimate. The detected point is shown with a red dot.}
\label{fig:template-matching}
\end{figure}

It should be noted that the prediction obtained from this method for localization is improved with sub-pixel accuracy. For this purpose, the search around the same point is performed again, this time by increasing the spatial dimensions of the target and template images. Finally, depth is computed via triangulation using the disparity (horizontal offset between matched points) as per Eq. \ref{eq:depth}\cite{forsyth2002image}. In this equation, $b$ is the camera baseline, $f$ is the focal length, and $disparity$ is the difference in distance between the corresponding points in the images. Figure \ref{fig:template-matching} shows an overview of the matching process with this method.

\begin{equation}
depth = \frac{b \times f}{disparity}
\label{eq:depth}
\end{equation}

\subsubsection{Deep learning-based Stereo Depth Estimation}

In this approach, stereo images are fed into the object detection model. The convolutional neural network in the first module estimates the probable locations of flower centers in both images. Subsequently, another convolutional neural network receives the resulting probability maps, along with the intermediate features of the input images from the encoder, to estimate the depth of the flower centers relative to the left camera. It is important to note that the matching occurs implicitly, and the depth is directly predicted by the network.

\begin{figure}[H]
\centering
\includegraphics[width=0.6\textwidth]{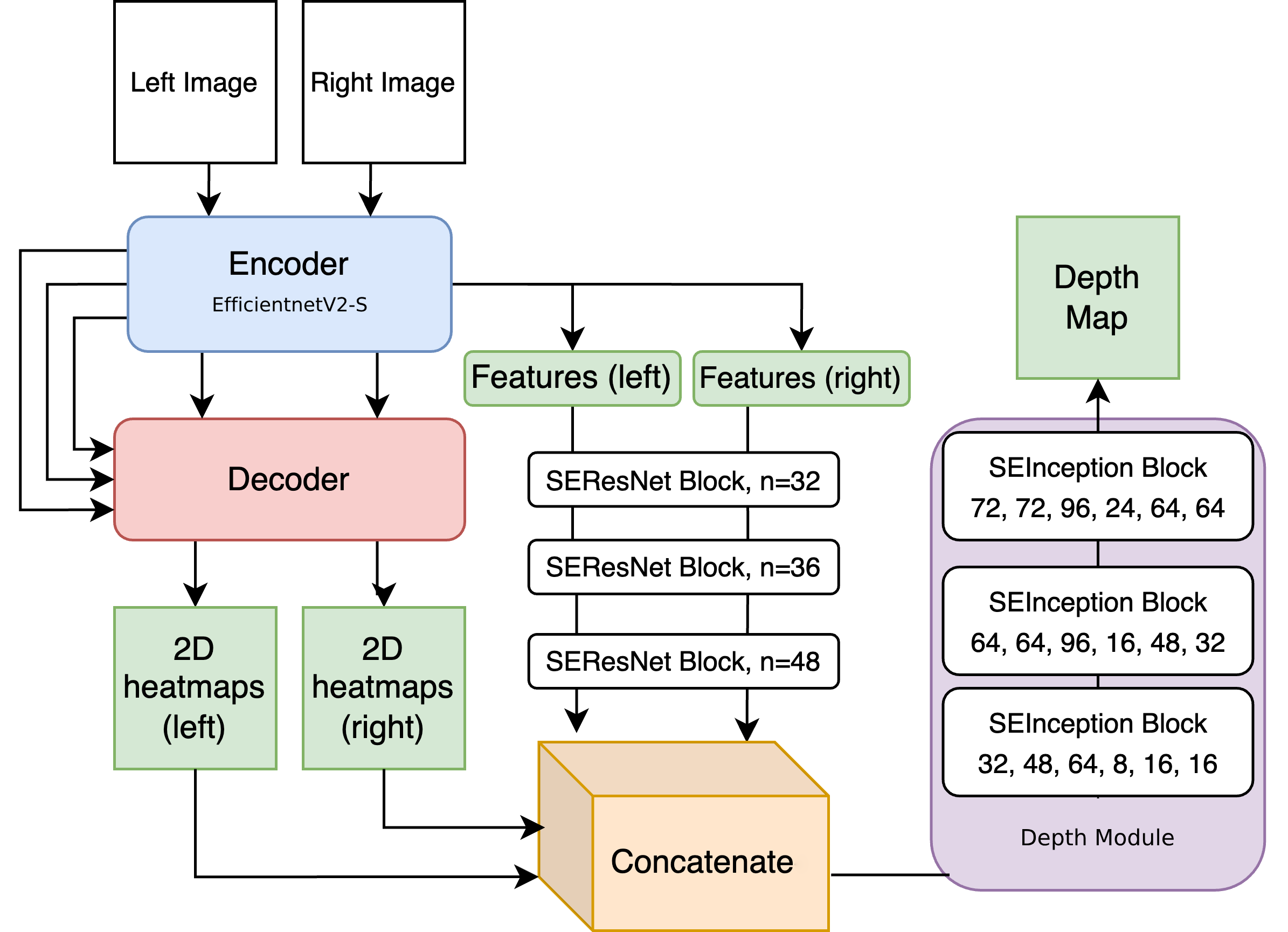}
\caption{Structure of the proposed deep learning-based model for the stereo method. The decoder is the same as that in monocular model.}
\label{fig:stereo-structure}
\end{figure}

The structure of the depth estimation network is shown in Figure \ref{fig:stereo-structure}. The proposed model consists of an encoder and two decoders. The encoder uses pre-trained weights on the ImageNet dataset and is responsible for extracting features from the input stereo images. The weights of this network are shared for both input images. The two decoder modules are each responsible for 2D localization and scene depth estimation, respectively. These two decoder modules have skip connections to the encoder layers.

The output of the 2D localization has three channels for each image, adding up to six channels. These three channels include: the probability map of near flowers, distant flowers, and the background. These heatmaps along with extracted features of both images are then fed to the depth estimation network which directly predicts a dense heatmap for the scene. The depth of flower centers is further extracted from this map.

\subsubsection{Loss Function}
The loss function is a combination of two loss functions for correct classification of pixels (for flower center detection) and the depth estimation loss function. The formula of the total loss function is defined as follows:

\begin{equation}
L_{total} = \sum_{i \in \{\text{right}, \text{left}\}} L_{\text{CCE}}(i) + L_{Depth}
\end{equation}

where $L_{\text{CCE}}$ is the categorical cross-entropy loss as previously defined, and $L_{\text{Depth}}$ is the smoothed L1 loss, also as previously defined.

\section{Experimental Results}
In this section, the training setup and process for both monocular and stereo localization methods are described, followed by a comprehensive presentation of the results. The results are organized into two main parts: the first addresses 2D detection and localization of flowers, while the second reports findings on depth estimation. For each part, multiple approaches were implemented and their respective results are provided. Notably, all models were trained exclusively with synthetic data, as depth information was unavailable for real-world data.

\subsection{Training Setup}

The proposed models are trained for $25$ epochs with a learning rate of $0.001$ and a batch size of $4$. During training, the error on the validation dataset is monitored to prevent overfitting. The model training can be performed in either one stage, where both localization and depth estimation modules are trained jointly, or in two stages, where the localization module is trained first. Then after freezing the localization module's weights, the depth estimation module is trained. Based on the experiments, the two-stage training approach achieved better results.

\subsubsection{Monocular Training Process}
\begin{figure}[H]
\centering
\includegraphics[width=0.8\textwidth]{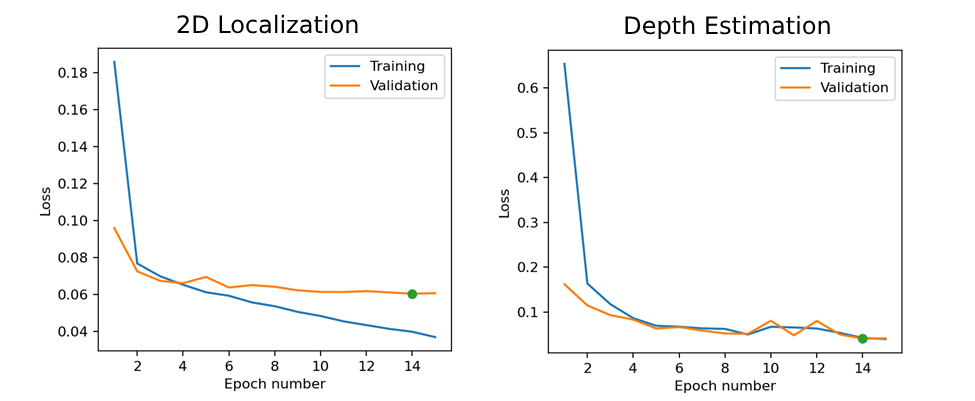}
\caption{2D localization and depth estimation loss values during model training. The green point indicates the model with the lowest validation error.}
\label{fig:detection-and-depth-loss}
\end{figure}

Figure \ref{fig:detection-and-depth-loss} shows the loss values for the 2D object localization and depth estimation modules during training. The loss values are plotted for the training and validation data.

\subsubsection{Stereo training Process}
\begin{figure}[H]
\centering
\includegraphics[width=0.8\textwidth]{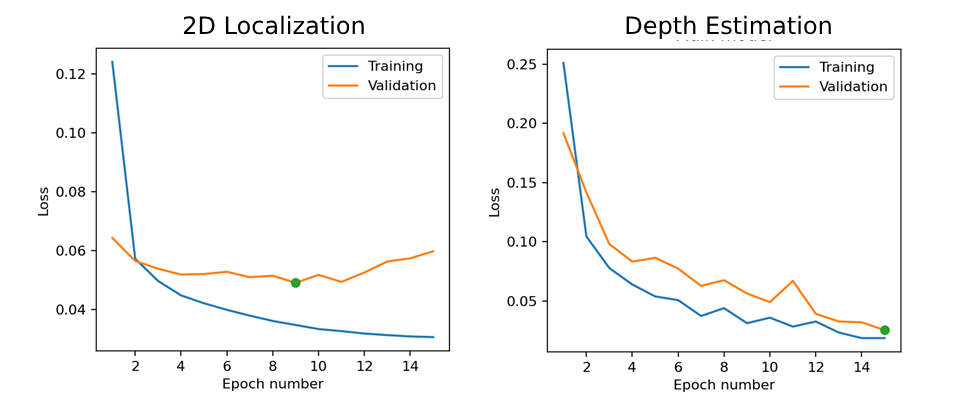}
\caption{2D localization and depth estimation loss values during model training. The green point indicates the model with the lowest validation error.}
\label{fig:stereo-detection-and-depth-loss}
\end{figure}

Figure \ref{fig:stereo-detection-and-depth-loss} shows the loss values for the object localization and depth estimation modules during training. The loss values are plotted for the training and validation data.

\subsection{Evaluation Metrics}
For object detection, precision, recall, and F-score are computed using a spatially relaxed criterion. A true positive (TP) is registered if a prediction exceeds the confidence threshold (0.51) within a ±5-pixel window around a ground-truth point. This window threshold is equivalent to setting an Intersection over Union (IoU) threshold for bounding box overlap in anchor-based object detection methods. In such methods, if the overlap between the predicted bounding box and the ground truth exceeds a certain threshold (e.g., 50\% or higher), the prediction is deemed correct. Conversely, a predicted point is a false positive (FP) if no ground truth exists within its window. False negatives (FN) occur when no prediction is made near a ground-truth point.

For depth estimation, mean absolute error (MAE) is calculated on valid pixels (ground-truth depth$> 0$). The implementation includes a transformation of the sigmoid-output predictions back to the metric depth space, followed by MAE computation between aligned prediction and ground-truth pixels.

\subsection{Monocular 2D Localization of Flowers}
Table \ref{tab:2d-localization-results} shows the results of 2D localization of flower centers using monocular images on the test set of synthetic and real datasets, respectively. The term "Near Flower" refers to a flower located within a distance of 2 meters.

\begin{table}[H]
\centering
\caption{2D localization results on both synthetic and real test data. The model is trained on the synthetic dataset, given a single image as input.}
\label{tab:2d-localization-results}
\begin{tabular}{@{}lcccccc@{}}
\toprule
& \multicolumn{3}{c}{Synthetic Data} & \multicolumn{3}{c}{Real Data} \\
Category & Precision & Recall & F-Score & Precision & Recall & F-Score \\ \midrule
Near Flower & 94.4 & 100 & \textbf{96.6} & 81.9 & 74.6 & 78.0 \\
Distant Flower & 94.3 & 100 & \textbf{96.3} & 60.2 & 80.5 & 68.8 \\
\bottomrule
\end{tabular}
\end{table}

\subsection{Stereo 2D Localization of Flowers}
Table \ref{tab:stereo_2d_localization} displays the results for 2D localization of flower centers given stereo images as input on both the synthetic and real datasets.

\begin{table}[H]
\centering
\caption{2D localization results given stereo images. The results are reported for the test set of both synthetic and real datasets.}
\label{tab:stereo_2d_localization}
\begin{tabular}{lcccccc}
\toprule
& \multicolumn{3}{c}{Synthetic Dataset} & \multicolumn{3}{c}{Real Dataset} \\
Category & Precision & Recall & F-Score & Precision & Recall & F-Score \\ \midrule
Near Flower & 94.4 & 97.6 & 95.5 & 76.2 & 72.7 & 74.4 \\
Distant Flower & 99.6 & 100 & 99.8 & 69.1 & 63.2 & 66.0 \\
\bottomrule
\end{tabular}
\end{table}

\subsection{Monocular Depth Estimation}
Table \ref{tab:monocular_depth_estimation_error} displays the L1 error of the depth estimation module in monocular scenario.

\begin{table}[H]
\centering
\caption{L1 Error of the monocular depth estimation module for the synthetic data.}
\label{tab:monocular_depth_estimation_error}
\begin{tabular}{lc}
\hline
Category & L1 Error (meters) \\
\hline
Close Flower & 0.16 \\
Distant Flower & 0.23 \\
\hline
\end{tabular}
\end{table}

\begin{figure}[H]
  \centering
  \includegraphics[width=\linewidth]{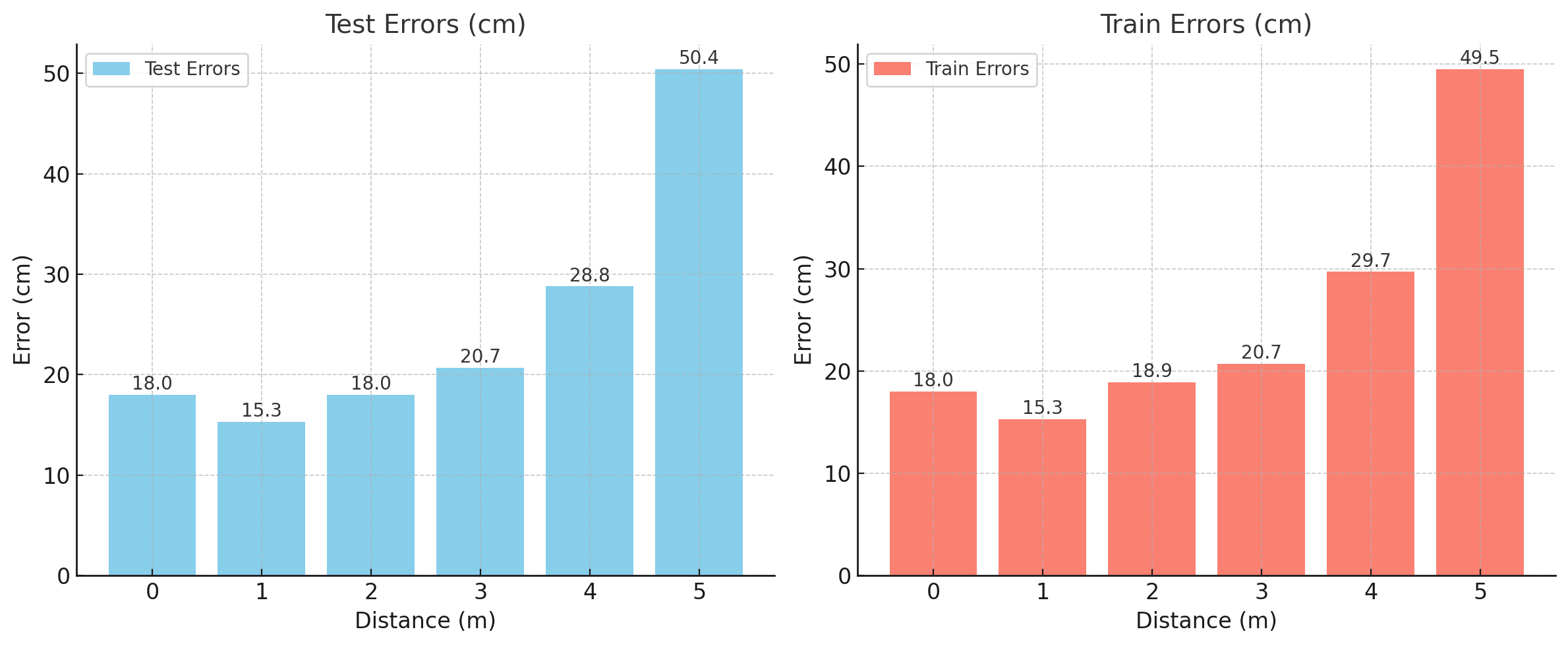}
  \caption{(left): The training L1 error of the monocular depth estimation module, categorized by distance ranges from the camera. (right): The error on the test data. Distances are in meters and errors are in centimeters.}
  \label{fig:l1_error_monocular_depth}
\end{figure}

Figure \ref{fig:l1_error_monocular_depth} shows the L1 error rate for various depth ranges on both training and test subsets of the synthetic data.

\subsection{Stereo Depth Estimation with Template Matching}
This study employs template matching (as an alternative) for stereo depth estimation, constrained by epipolar geometry to limit template search space. Two correlation criteria were evaluated, with results detailed in subsequent sections. Since flower positions derive directly from the single-image 2D localization model, localization errors are similar to those of the monocular approach. Depth estimation errors for this method are quantified in Table \ref{tab:stereo_depth_estimation_error_template_matching}.

\begin{table}[H]
\centering
\caption{Stereo depth estimation errors (in meters) on synthetic data  using template matching algorithm. NCCoef and NCCorr refer to the criteria used for the template matching algorithm.}
\label{tab:stereo_depth_estimation_error_template_matching}
\begin{tabularx}{\textwidth}{lXXXXX}
\toprule
Subset & Category & L1 Error (NCCoef) & L1 Error (NCCorr) \\
\midrule
Train & Near Flower & 0.1 & 0.12 \\
 & Distant Flower & 0.2 &  0.2  \\
 & Overall & 0.17 &  0.18  \\
\midrule
Test & Near Flower & 0.06 & 0.09 \\
 & Distant Flower & 0.2 & 0.2  \\
 & Overall & 0.16 &  0.17  \\
\bottomrule
\end{tabularx}
\end{table}

\begin{figure}[H]
  \centering
  \includegraphics[width=0.8\linewidth]{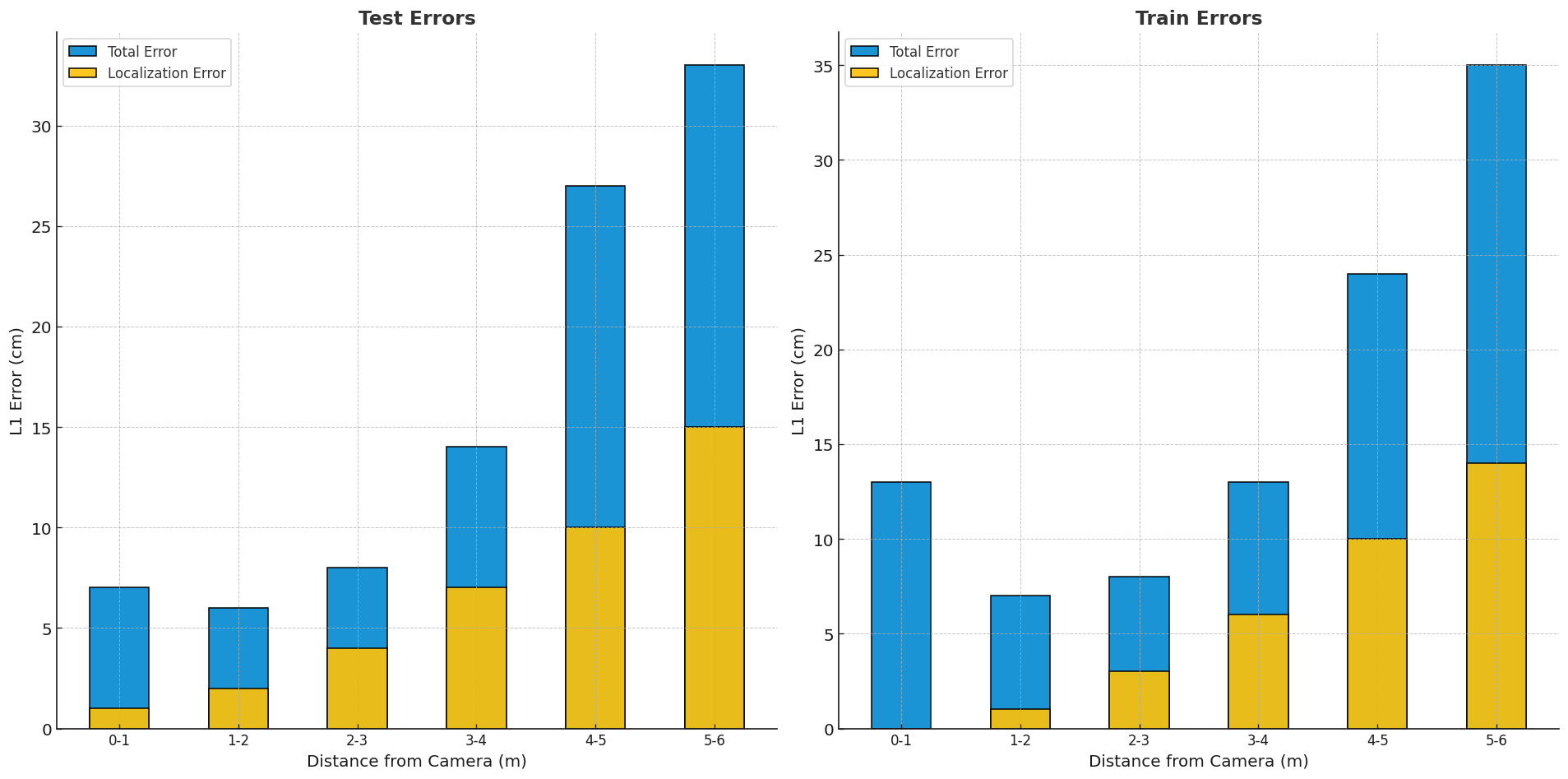}
  \caption{The training and test L1 errors of the stereo depth estimation module using template matching and triangulation, categorized by distance ranges from the camera. Distances are in meters.}
  \label{fig:l1_error_stereo_depth_tm}
\end{figure}

The results show no significant difference in performance between the two criteria. Figure \ref{fig:l1_error_stereo_depth_tm} illustrates the estimation error for various depth ranges using the NCCoef criterion. The yellow section indicates the contribution of the object localization module to the total error. To compute this, 2D localization predictions are replaced with ground-truth labels, the depth estimation algorithm is applied, and the error is measured. Subtracting this value from the actual total depth estimation error reveals the localization module’s contribution.

\subsection{Deep Learning Based Stereo Depth Estimation}

Table \ref{tab:stereo_depth_estimation_error_tm_synthetic} quantifies the L1 depth estimation error of our deep-learning-based stereo module on synthetic datasets. Real-world performance metrics are excluded due to the absence of acquired depth ground-truth for real-world data.

\begin{table}[H]
\centering
\caption{L1 error of the proposed deep learning based stereo depth estimation module on synthetic data.}
\label{tab:stereo_depth_estimation_error_tm_synthetic}
\begin{tabular}{lcc}
\hline
Category & L1 Error (meters) \\
\hline
Near Flower & 0.096 \\
Distant Flower & 0.13 \\
\hline
\end{tabular}
\end{table}

\begin{figure}[H]
  \centering
  \includegraphics[width=\linewidth]{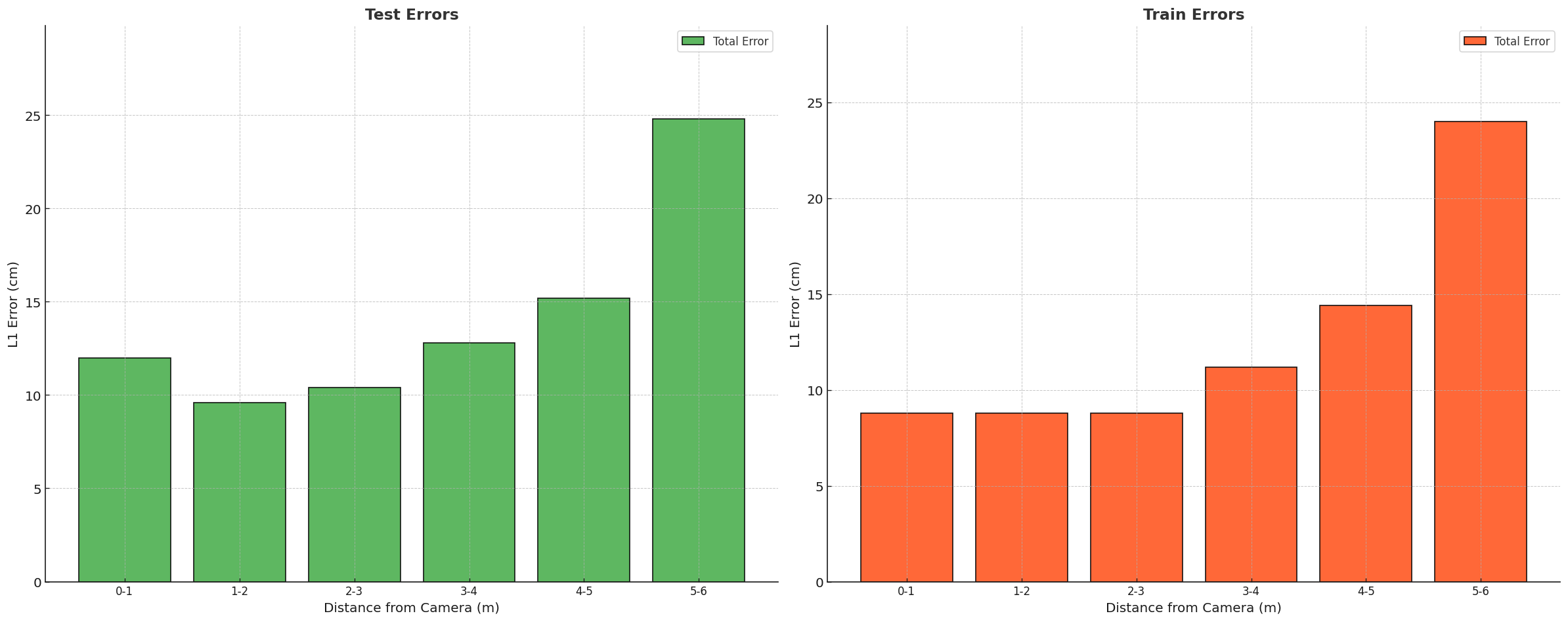}
  \caption{The L1 error of the proposed deep stereo depth estimation module for various depth ranges of train and test subsets. The errors are in centimeters.}
  \label{fig:stereo_depth_deep_results}
\end{figure}

The L1 error for various depth ranges is provided for both train and test subsets of the synthetic dataset in Figure \ref{fig:stereo_depth_deep_results}.

\subsection{Model Details and Inference Times}
Table \ref{table:combined-model-stats} presents a comparison of various deep learning models used for 2D localization and depth estimation, including their parameter counts, memory requirements, and inference times on both CPU and GPU. The models evaluated include monocular 2D localization, stereo 2D localization, monocular 3D localization, and stereo 3D localization. The GPU used was Nvidia Tesla P100 and the CPU was Intel Core i7 10810U.

\begin{table}[H]
\centering
\caption{Comparison of model parameters, memory requirements, and inference times across different models for localization and depth estimation. The L1 depth error is reported for near flowers.}
\label{table:combined-model-stats}
\begin{tabularx}{\textwidth}{p{3cm}XXXX}
\toprule
Localization Method & Parameters (Million) & Memory (MB) & Inference Time (CPU/GPU) (ms) & L1 Depth Error \\
\midrule
3D Monocular & 5.4  & 23.4   & 211/893  & 0.16 \\
Stereo 3D with Template Matching   & 3.8  & 19.5 & -/650  & 0.06 \\
3D Stereo & 4.7  & 23.8 & 270/1120  & 0.096 \\
\bottomrule
\end{tabularx}
\end{table} 

\section{Discussion}
In this section, a comprehensive analysis of both quantitative and qualitative results is provided. The quantitative results are also compared to similar studies in the agriculture field.

\subsection{Analysis, Interpretation and comparison of Quantitative Results}
Table \ref{tab:summary_results_synthetic} provides a summary of quantitative results for the test subset of the synthetic dataset.

\begin{table}[H]
\centering
\caption{Summary of results of 2D localization and depth estimation of the proposed methods on synthetic data.}
\label{tab:summary_results_synthetic}
\begin{tabular}{p{3cm}ccccp{2cm}}
\hline
Model Name & Category & Precision & Recall & F-score & Depth L1 Error \\
\hline
Deep Monocular & Near Flower & 94.4 & 100 & 96.6 & 0.16 \\
& Distant Flower & 94.3 & 100 & 96.3 & 0.23 \\
Deep Stereo & Near Flower & 94.4 & 97.6 & 95.5 & 0.096\\
& Distant Flower & 99.6 & 100 & 99.8 & 0.13 \\
Template Matching (NCCoef) Stereo & Near Flower & 94.4 & 97.6 & 95.5 & 0.06 \\
& Distant Flower & 99.6 & 100 & 99.8 & 0.2 \\
\hline
\end{tabular}
\end{table}

The quantitative results from the synthetic dataset indicate that both stereo and monocular image localization methods achieve relatively high and acceptable accuracy levels. The recall metric for both approaches indicates near-perfect localization of flowers, with recall rates of approximately 100\%. However, the precision metric is lower due to the presence of false positive predictions. This discrepancy can be attributed to the nature of the proposed solution, which generates dense predictions for all pixels, resulting in false positives. Adjusting the decision threshold in the probability map of flower centers can help mitigate the issue of false positive predictions.

Regarding the results of depth estimation, stereo depth estimation methods outperform monocular ones, consistent with the benchmark results of autonomous driving research \cite{6248074}. This is due to the increased spatial information available. Notably, the depth estimation error is higher for distant flowers, likely due to their increased distance from the camera, resulting in fewer pixels being captured in the image and consequently increasing the likelihood of model errors.

\begin{figure}[H]
  \centering
  \includegraphics[width=0.7\linewidth]{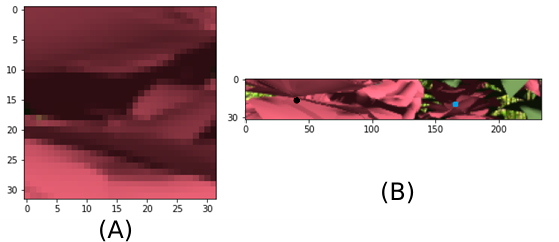}
  \caption{An example of the template matching error. The black point indicates the algorithm's detection, and the blue point indicates the label. (A) enlarged template (B) search window.}
  \label{fig:example_image_for_tm_error_behaviour_over_ranges}
\end{figure}

The template matching method for depth estimation demonstrates a better performance for near flowers compared to deep learning (5cm vs 8cm). However, it performs worse for distant flowers (18 cm vs. 13 cm). A detailed analysis reveals that the error is initially higher within 0 to 1 meter, then decreases before increasing again with distance. This fluctuation may result from the assumption of a constant template size, leading to reduced matching accuracy, especially for flowers within the 0- to 1-meter range, where the template captures limited visual information, as shown in Figure \ref{fig:example_image_for_tm_error_behaviour_over_ranges}.

At greater distances, small disparities mean that even a single-pixel error in the template matching module can lead to significant depth errors. 

According to Table \ref{tab:2d-localization-results}, the quantitative results of 2D detection models on real data are weaker than those obtained for synthetic data. This is mainly due to the fact that the deep learning models are trained on synthetic data, where many real-world conditions are not present. Such conditions include blur in images due to camera movement or lack of lens focus, environmental noise, and the limited variability of the 3D model used for flower bushes, which are considered weaknesses of the generated synthetic data.

Table \ref{tab:comparison_similar_studies} displays the hit rate (recall) of detection and depth estimation error in percentage for similar studies. The reported values for the proposed method of this research is related to near flowers of the test subset.

\begin{table}[htbp]
\centering
\caption{Performance comparison of the proposed method with similar methods on localization errors and depth estimation errors in studies related to agricultural products.}
\label{tab:comparison_similar_studies}
\resizebox{\textwidth}{!}{
\begin{tabular}{p{2cm}p{2cm}p{1cm}p{3cm}p{3cm}p{1.5cm}p{3cm}}
\hline
Study & Product & Image Count & Localization Method & Depth Estimation Method & Detection Hit Rate & Depth Estimation Error \\ \hline
\cite{si2015location} & Apple & 160 & Circle detection & Correspondence Matching and Triangulation & 89.5\% & 5\% in 40cm \& 1\% in 150cm \\ \hline
\cite{sun2021apple} & Apple Blossom & 165 & Segmentation using DeepLab-ResNet & - & 89\% &  \\ \hline
\cite{sun2021apple} & Peach Blossom & 24 & Circle detection & - & 78.0\% &  \\ \hline
\cite{sun2021apple} & Pear Blossom & 18 & Circle detection & - & 85.4\% &  \\ \hline
\cite{hiary2018flower} & Flowers & 8189 (oxford102) &  VGG16 Encoder-Decoder based Segmentation & - & 85.9\% &  \\ \hline
\cite{hiary2018flower} & Flowers & 1360 (oxford17) & VGG16 Encoder-Decoder based Segmentation & - & 82.2\% &  \\ \hline
\cite{prakash2022flower} & Flowers & 1360 (oxford17) & Yolov4 & - & 98.0\% &  \\ \hline
\cite{li2022real} & Kiwi Blossom & 1451 & Yolov4 & - &  97.6\% (Yolov4) &  \\ \hline
\cite{WANG2022106716} & Apple & 768 & A3N Network & A3N &  87.3\% & 0.61cm \\ \hline
\cite{chen2024rapidstraw} & Strawberries & 4860 & Yolov8 & Enhanced Yolov8-Pose &  97.85\% (mAP) & 1.16\% \\ \hline
\cite{JANG2024108961} & Tomato & 1530 & Yolov8 & Direct depth acquisition &  96.30\% $AP_{50}$ & - \\ \hline
\cite{kohan2011robotic} & Damask Rose & 30 & Thresholding & Correspondence Matching and Triangulation &  98\% & 2\% up to 100cm \\ \hline
Ours (Stereo) & Damask Rose & 1000 & Deep Learning & Template Matching and Triangulation &  100\% & 3\% up to 200cm \\ \hline
Ours (Stereo) & Damask Rose & 1000 & Deep Learning & Deep Learning &  100\% & 5\% up to 200cm \\ \hline
Ours \small{(Monocular)} & Damask Rose & 1000 & Deep Learning & Deep Learning &  100\% & 8\% up to 200cm\\ \hline
\end{tabular}%
}
\end{table}

\begin{figure}[H]
  \centering
  \includegraphics[width=0.5\linewidth]{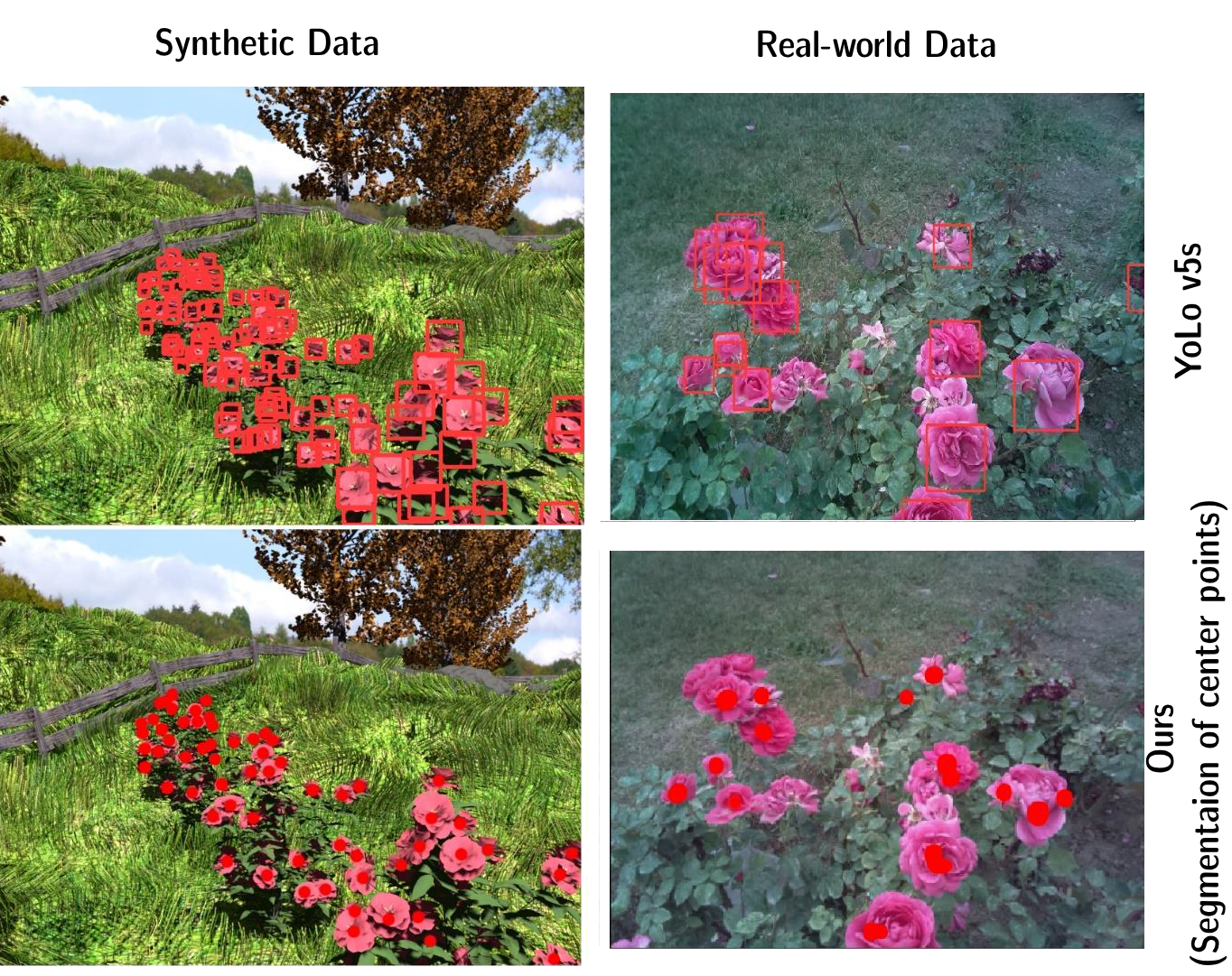}
  \caption{Qualitative comparison of rose detection on synthetic (left) and real-world (right) samples using YOLOv5s (top) and our segmentation-based method (bottom). YOLOv5s shows higher accuracy on real-world data, while our method performs better on synthetic data.}
\label{fig:qualititative-comparison-with-yolo}
\end{figure}

For a fair comparison, a YOLOv5s model, (the small version of YOLOv5 with 7 million parameters), was fine-tuned on the synthetic dataset. It achieved an 83\% detection accuracy for roses in the test dataset, which is 17\% lower than the proposed model. Since only the rose center coordinates were labeled, bounding box dimensions were programmatically generated using the heuristic formula:
\begin{equation}
    \text{Bounding box size (pixles)} = \frac{60}{\text{depth (meters)}}.
\end{equation}
Predictions with an Intersection over Union (IoU) overlap greater than $0.5$ were considered correct. On real-world data, YOLOv5s achieved $4\%$ higher detection accuracy than the proposed method, indicating better generalization despite underperforming on synthetic data (see Figure\ref{fig:qualititative-comparison-with-yolo}). Table \ref{tab:2d_localization_comparison_with_yolo} compares these results.

\begin{table}[htbp]
\centering
\caption{Performance comparison of the proposed point-based 2D localization methods with fine-tuned YoLov5s on realistic dataset for near flowers.}
\label{tab:2d_localization_comparison_with_yolo}
\begin{tabular}{lccc}
\hline
Method & Precision & Recall & F-Score\\
\hline
Monocular Point-based & 81.9 & 74.6 & 78.0 \\
Stereo Point-based & 76.2 & 72.7 & 74.0 \\
Stereo Yolov5s & 82 & 78.8 & 80.4 \\
\hline
\end{tabular}
\end{table}

Based on the comparison table, the proposed method for localizing damask roses has achieved a 100\% hit rate on synthetic data, which is 2\% higher compared to \cite{kohan2011robotic}. Furthermore, when compared to recent works in the field of localizing other flowers, it shows 2 to 3\% higher hit rates than methods using Yolov4. Compared to works \cite{hiary2018flower} and \cite{sun2021apple}, the hit rate of the proposed method has been significantly higher. However, since the model is only trained on synthetic data, the detection hit rate is lower on real data compared to other algorithms. 

In the depth estimation section, the proposed method using template matching achieved an error rate of 3\% for a distance range of 2 meters, which is competitive with \cite{kohan2011robotic}. Furthermore, the deep learning-based method has an error rate of 6.5\% in the range of 2 to 6 meters, demonstrating good performance considering the larger and more challenging estimation range. The monocular deep learning-based method, as expected, results in the highest depth estimation error.

\subsection{Qualitative Analysis of Results}
In this section, the qualitative results obtained are provided. The analysis of the results is presented at the end of this section.

\subsubsection{Single-image Method}
\begin{figure}[H]
  \centering
  \includegraphics[width=0.7\linewidth]{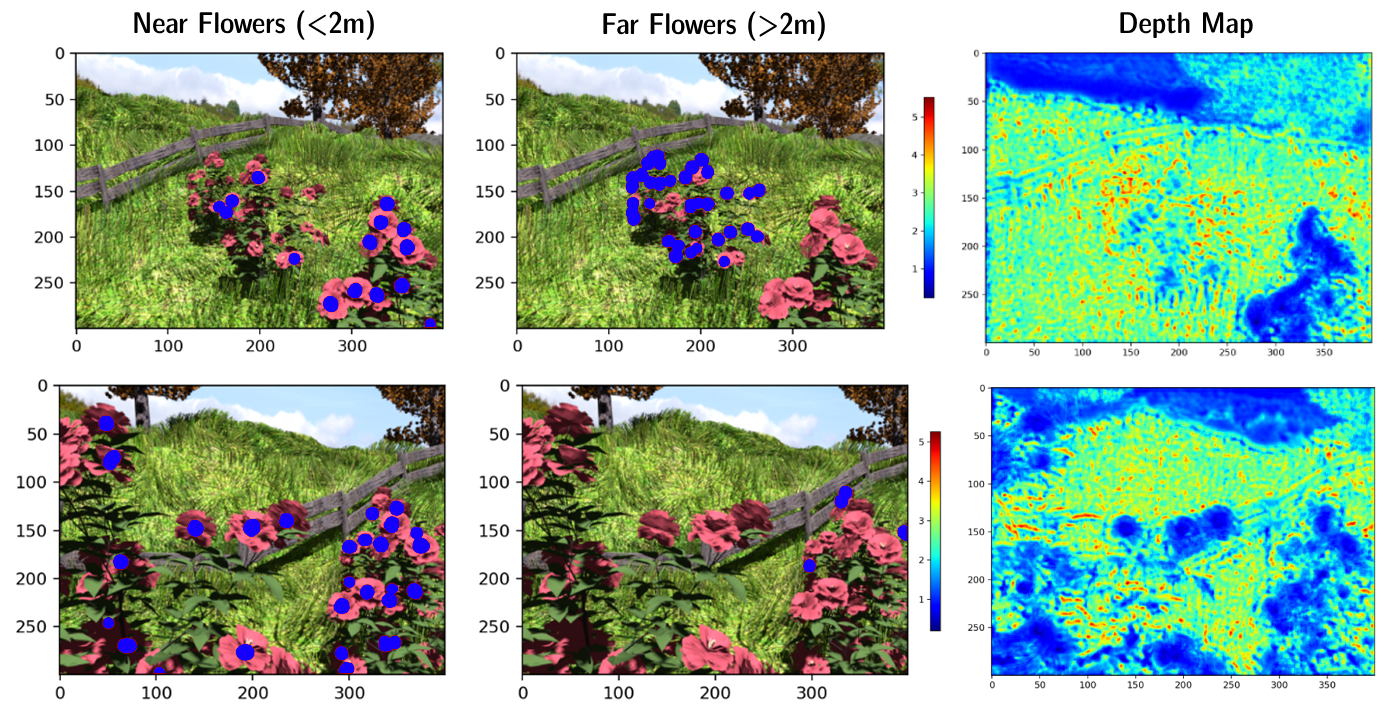}
  \caption{Monocular deep learning model results on synthetic data: 2D localization and depth estimation for near and distant flowers (left to right). In the depth maps, depth estimation accuracy is limited to flowers; other objects and surfaces yield unreliable results.}
\label{fig:monocular_localization_sample}
\end{figure}

The qualitative results of flower localization are discussed in this section. Figure \ref{fig:monocular_localization_sample} displays two sample plots of monocular 2D localization and depth estimation on the synthetic dataset. In each sample, flowers localized in the near category (within 2m distance from the camera) are presented on the left side, while flowers in the distant category (over 2m) are shown on the right side. Below each detection, the corresponding depth estimation is also provided. 

\begin{figure}[H]
  \centering
  \includegraphics[width=0.7\linewidth]{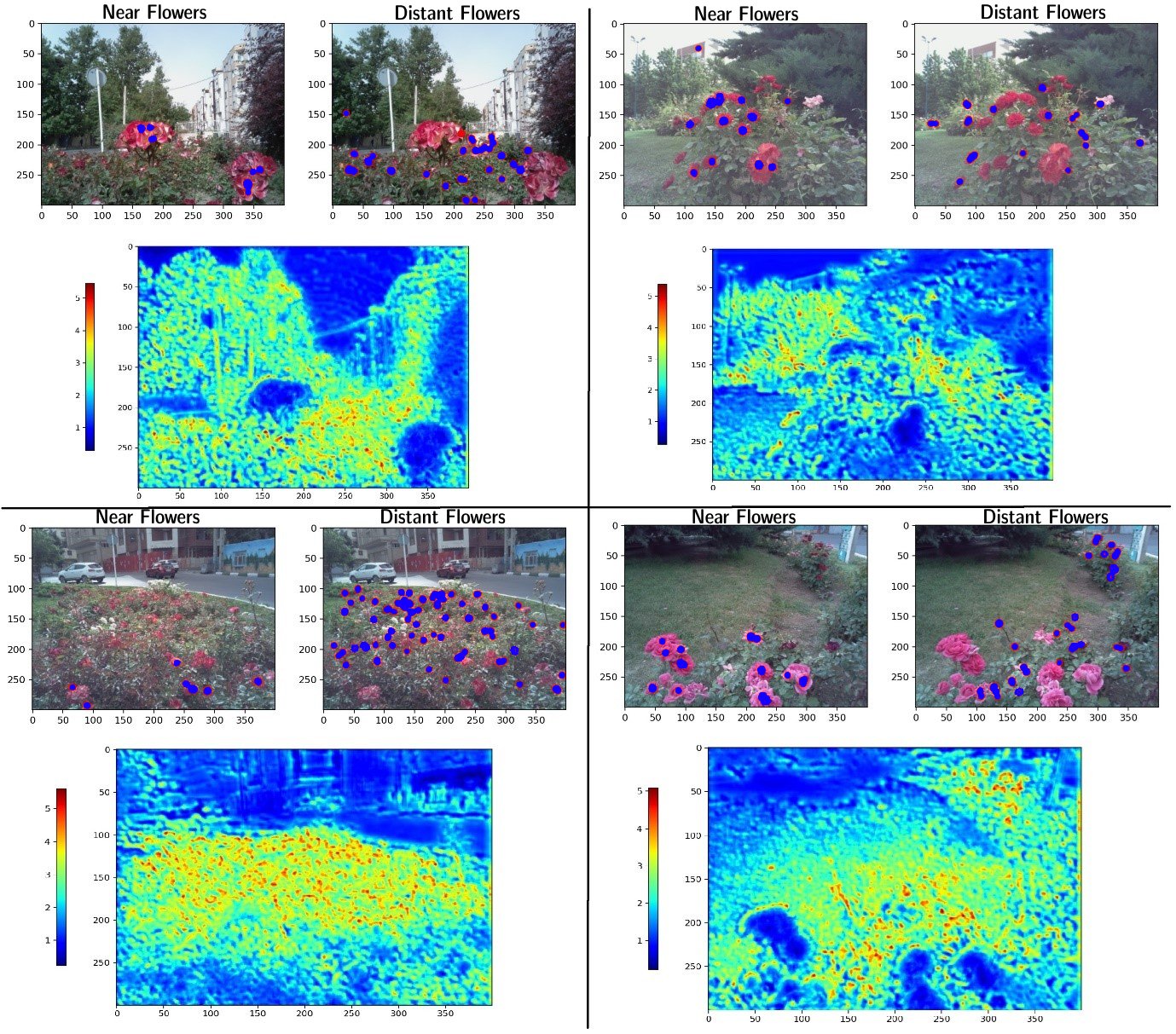}
  \caption{Four examples of monocular 2D localization and depth estimation results on real data. In each sample, the top-left image shows near-flower and the top-right shows distant-flower localization results. In the depth maps, depth estimation accuracy is limited to flowers; other objects and surfaces yield unreliable results.}
  \label{fig:examples_of_monocular_method_on_real_data}
\end{figure}

Figure \ref{fig:examples_of_monocular_method_on_real_data} displays four examples of monocular 2D localization and depth estimation results on real data.

\subsubsection{Stereo-based Deep Learning Method}
\begin{figure}[H]
  \centering
  \includegraphics[width=0.75\linewidth]{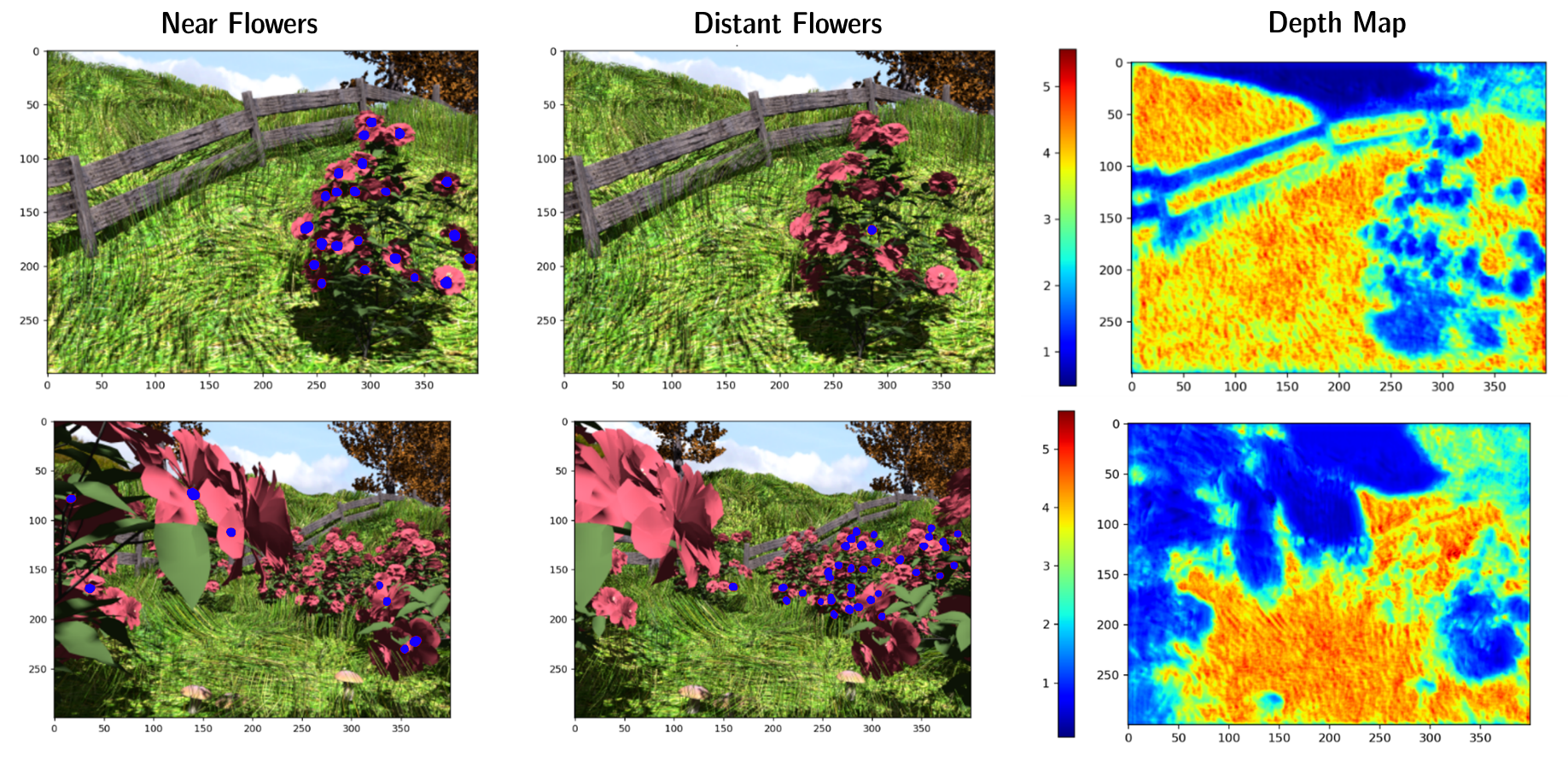}
  \caption{Stereo-based deep learning model results on synthetic data: 2D localization and depth estimation for near and distant flowers (left to right). Depth maps show accurate estimation only for flowers.}
\label{fig:example_stereo_method}
\end{figure}

\begin{figure}[H]
  \centering
  \includegraphics[width=\linewidth]{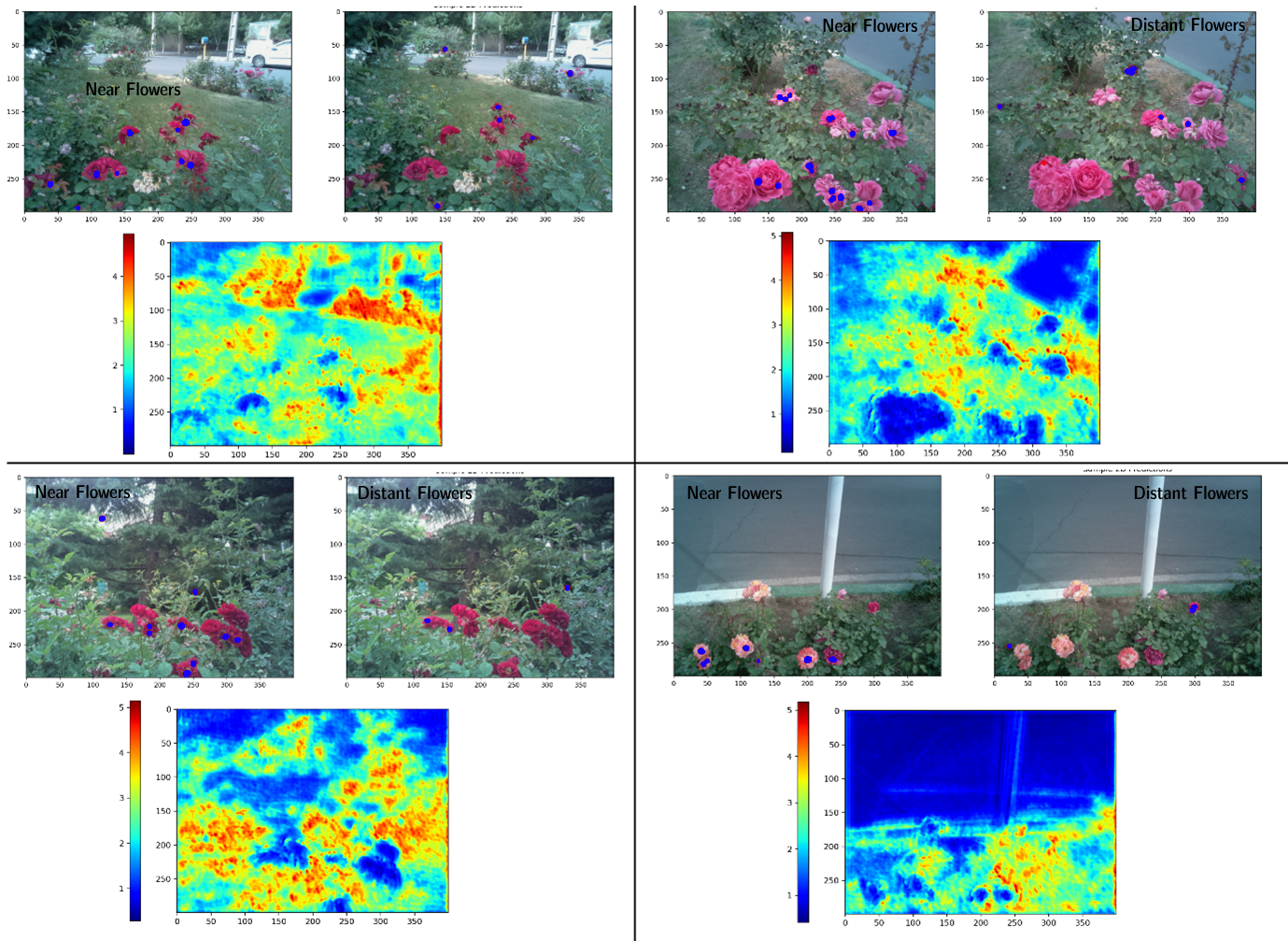}
  \caption{Four real-data examples of 2D detection (near/distant flowers) and depth estimation. Depth maps are reliable only for flowers.}
\label{fig:example_images_deep_stereo_real_data}
\end{figure}

Figure \ref{fig:example_stereo_method} shows two examples of 2D localization and depth estimation results for roses on synthetic data. Figure \ref{fig:example_images_deep_stereo_real_data} also displays four localization and depth estimation on real data.

\subsubsection{Stereo Depth Estimation Method Based on Template Matching Algorithm}

\begin{figure}[H]
  \centering
  \includegraphics[width=0.6\linewidth]{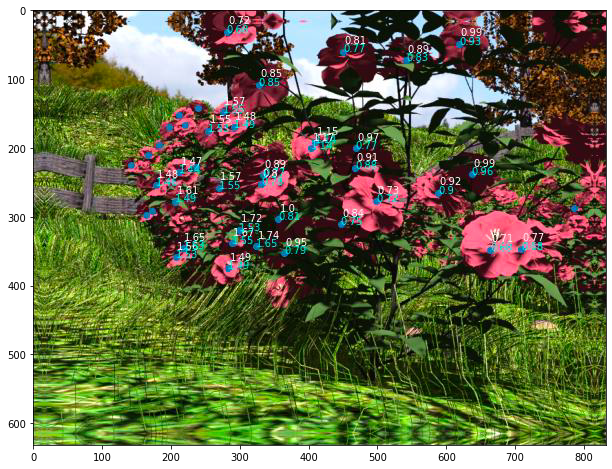}
  \caption{An example of depth estimation using the template matching method. White values indicate ground truth labels and blue ones indicate predictions. The blue dots indicate flower centers.}
  \label{fig:first_example_template_matching_predictions}
\end{figure}

Figure \ref{fig:first_example_template_matching_predictions} shows an example of the depth estimation results using the template matching method.

\subsubsection{In depth Analysis of Qualitative Results}
The deep learning-based methods show strong performance in localizing flowers in synthetic data, with the main issue being false positives, which can be mitigated by increasing the decision threshold. The qualitative results demonstrate that the model effectively distinguishes between near and distant flowers (less than 2 meters) in both left and right images. This accurate localization supports the depth estimation module, contributing to its accelerated learning process.

\begin{figure}[H]
  \centering
  \includegraphics[width=\linewidth]{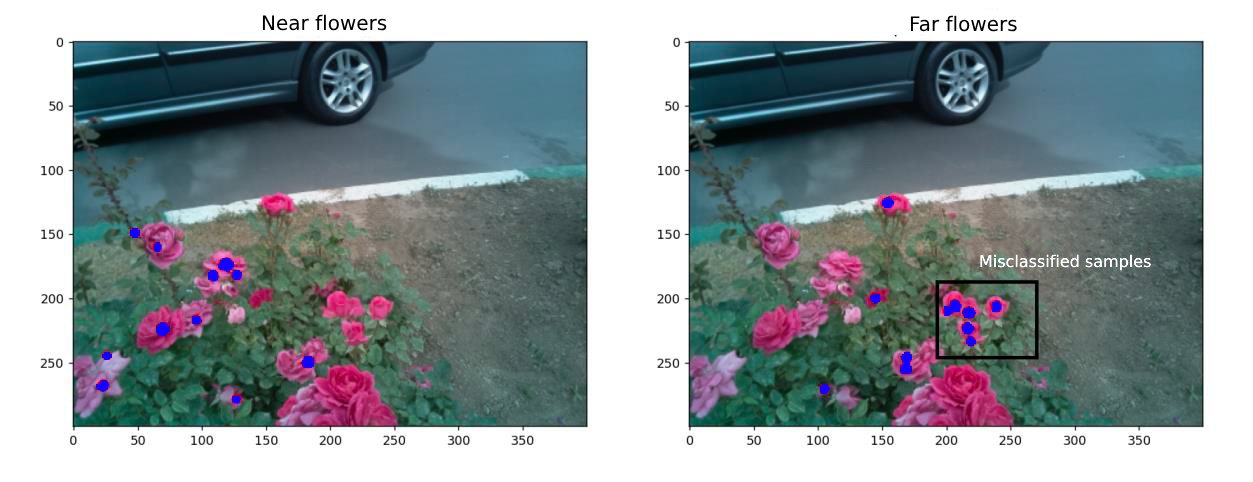}
  \caption{An example of misclassification of a flower in real data.}
  \label{fig:example_misclassification_of_near_flower}
\end{figure}

The qualitative analysis of 2D flower localization on real data reveals two main issues. First, decreased image registration quality at distant points, especially when the camera focuses on nearby objects, complicates the detection of distant flowers. Second, the model sometimes misclassifies nearby, small-sized flowers as distant ones. Figure \ref{fig:example_misclassification_of_near_flower} illustrates this problem, highlighting a weakness of synthetic data: the lack of size diversity among flowers at varying distances.

Based on the qualitative results of the depth estimation using deep learning, the depth estimations related to objects and surfaces other than flowers are not reliable. This is because the supervision of deep learning models is solely based on the centers of flowers. Moreover, significant discontinuities are observed in the depth predictions for such surfaces. This discontinuity is more prominent for single-image models. However, since the primary goal is to accurately estimate the depth of the flower centers, this does not pose a problem.

According to the first example prediction of the monocular model, a significant decrease in depth estimation accuracy is observed when estimating the depth of distant flowers. This estimation is much more accurate for stereo models, with a higher accuracy for distant flowers.

Based on figure \ref{fig:first_example_template_matching_predictions}, depth estimation using the template matching method demonstrates good results.

\begin{figure}[H]
  \centering
  \includegraphics[width=\linewidth]{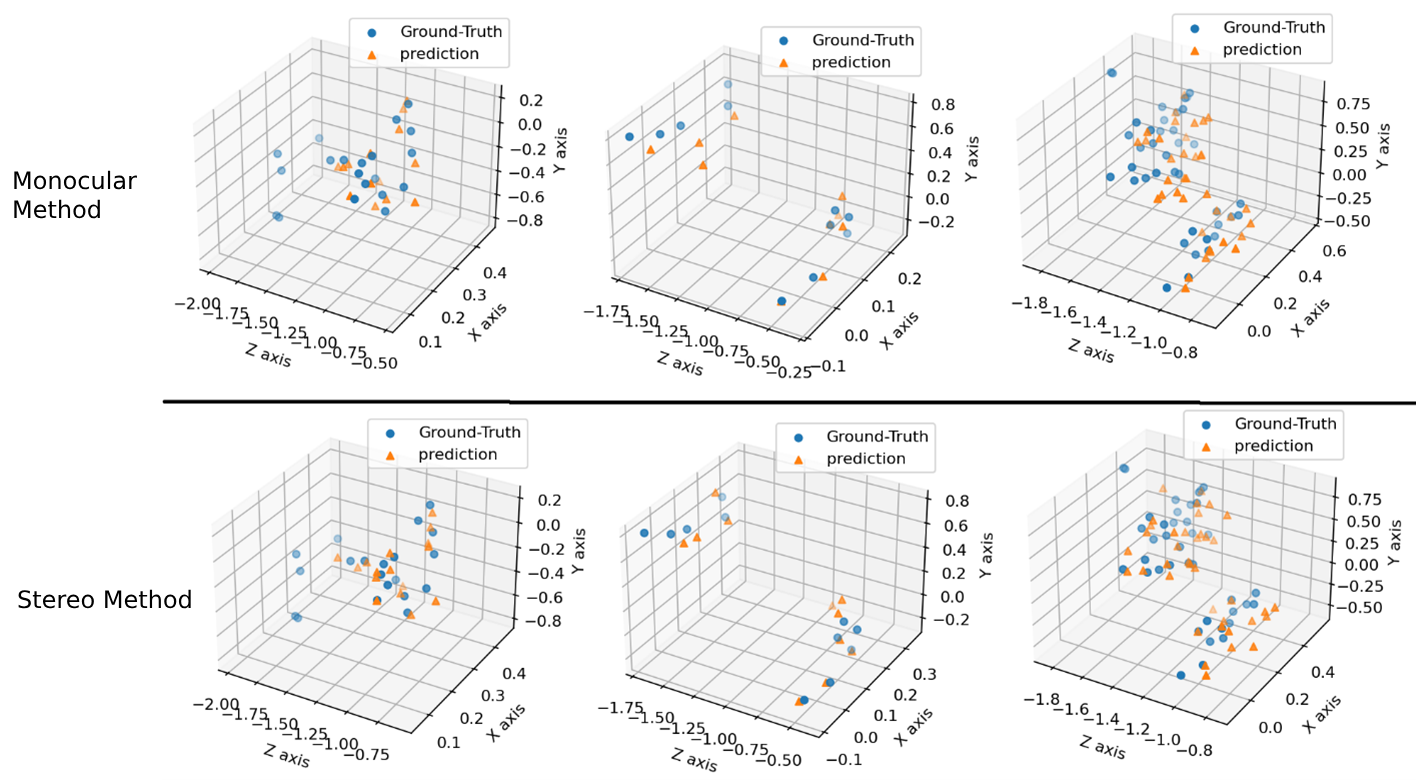}
  \caption{This plot shows the 3D coordinates of flower center points for three synthetic data samples. Circles represent the actual labels and triangles indicate the model predictions. The top section displays predictions from the monocular model, and the bottom section shows predictions from the stereo deep learning based model.}
  \label{fig:flower_center_points_in_3d}
\end{figure}

Figure \ref{fig:flower_center_points_in_3d} compares the performance of single-image and stereo deep learning algorithms. Three random synthetic data samples are selected, and the models' predictions are projected into 3D coordinates alongside the true values. The plot illustrates the superior performance of the stereo method, as the single-image model exhibits significant errors in predicting the depth of distant points.

\section{Conclusion and Future Research}
This study presents a 3D perception pipeline for robotic harvesting of Damask roses, addressing the critical challenge of labor-intensive manual harvesting through synthetic data-driven automation. By integrating a point-based object detection framework with stereo depth estimation, the proposed pipeline eliminates reliance on bounding boxes, simplifying implementation while achieving high 2D localization accuracy. Comparative analysis revealed that deep learning-based stereo depth estimation outperformed traditional template-matching approaches at long-range distances, though template matching demonstrated superior precision for close-to-camera flowers. These findings underscore the potential of synthetic pretraining to bridge domain gaps, reduce labeling costs, and enable robust real-world deployment of agricultural robots.

To further enhance the pipeline’s robustness and scalability, future research should prioritize expanding the diversity of the synthetic dataset by incorporating 3D rose models of varying sizes, growth stages, and occlusion scenarios. Such enhancements would address morphological disparities between synthetic and real data, thereby reducing depth estimation errors. Additionally, advanced data augmentation techniques—such as simulating lens distortion, motion blur, and dynamic environmental conditions—could improve model resilience under real-world imaging challenges.

Increasing the resolution of synthetic dataset images could improve matching accuracy, though it may impact execution speed. Implementing sub-pixel localization by predicting fractional deviations in x and y directions could increase the accuracy of depth estimation for deep learning-based methods. Utilizing deformable convolutional layers \cite{dai2017deformable}, which allow for adaptable sampling locations, could further boost localization accuracy.

Developing a criterion for adjusting template size based on the rose's distance from the camera could address matching errors associated with template size discrepancies. The assignment of the matching task to a neural network, such as a Siamese network \cite{koch2015siamese}, could improve the matching accuracy by operating at the feature level rather than the pixel level, as in research \cite{brandao2019widening}. 

To enhance depth estimation accuracy, the integration of self-supervised learning frameworks and dynamic feature fusion networks \cite{10043439} could help. Moreover, self-supervised approaches, such as those leveraging monocular or stereo sequences, reduce reliance on labeled data by enforcing geometric consistency constraints during training \cite{wang2020self}. Complementing this, dynamic feature fusion mechanisms enable adaptive weighting of multi-scale contextual features, improving robustness in textureless regions (e.g., rose petals) \cite{10043439}. Furthermore, transformer-based architectures offer superior global context modeling, addressing challenges in capturing long-range dependencies for precise localization \cite{9851497}. These techniques collectively address domain gaps and occlusion-related errors in agricultural robotics. Such ideas, combined with the proposed pipeline’s foundation, could significantly improve the adaptability and efficiency of robotic harvesting systems, paving the way for scalable automation in specialty crop agriculture.

\section{Declaration of generative AI and AI-assisted technologies in the writing process}

During the preparation of this work the author(s) used ChatGPT-4o's capabilities in order to improve the writing and grammar of some parts of the manuscript, since English was not the author's native language. The AI tool was not used to generate scientific content, analyze data, or create figures. After using this tool/service, the author(s) reviewed and edited the content as needed and take(s) full responsibility for the content of the publication.

\bibliographystyle{IEEEtran} 
\bibliography{cas-refs.bib}

\end{document}